\documentclass[11pt]{article}
\usepackage{acl}

\usepackage{times}
\usepackage{latexsym}
\usepackage[T1]{fontenc}
\usepackage[utf8]{inputenc}
\usepackage{microtype}
\usepackage{inconsolata}
\usepackage{booktabs}
\usepackage{amsmath,amssymb}
\usepackage{algorithm}
\usepackage{algpseudocode}
\usepackage{graphicx}
\usepackage{xcolor}
\usepackage{multirow}
\usepackage{subcaption}
\usepackage{colortbl}
\usepackage{pifont}
\usepackage{tikz}
\usetikzlibrary{shapes,arrows,positioning,fit,backgrounds}

\graphicspath{{figures/}}

\newcommand{\ael}{\textsc{AEL}}

\newcommand{\fcc}{\textsc{FCC}}
\newcommand{\llmfcc}{\textsc{LLM-FCC}}

\title{AEL: Evolving Agent Harness in Open-Ended Environments}

\author{
  \textbf{Wujiang Xu}\textsuperscript{1} \quad
  \textbf{Jiaojiao Han}\textsuperscript{1} \quad
  \textbf{Minghao Guo}\textsuperscript{1} \quad
  \textbf{Kai Mei}\textsuperscript{1} \\
  \textbf{Xi Zhu}\textsuperscript{1} \quad
  \textbf{Han Zhang}\textsuperscript{1} \quad
  \textbf{Dimitris N. Metaxas}\textsuperscript{1} \\
  \textsuperscript{1}Rutgers University
}

\begin{document}
\maketitle

\begin{abstract}
LLM Agents Harnesses are hand-designed and stay fixed, so agents accumulate experience but never learn \emph{how to use} it: which memories to retrieve, when retrieved evidence is misleading, and when the retrieval strategy itself should change. We introduce \emph{Agent Evolving Learning} (\ael{}), a two-timescale framework that evolves the harness, recasting memory use as online policy selection. A fast Thompson Sampling bandit selects among memory-retrieval policies episode by episode, while slow LLM reflection follows a \emph{diagnose-before-prescribe} principle: it first explains why performance degraded, then injects a targeted new retrieval policy as a bandit arm when the current pool plateaus. \ael{} outperforms ten self-improving and non-LLM baselines on a sequential portfolio benchmark, lifting Sharpe by 27\% over the strongest memory-only variant with the lowest variance among all stochastic methods, and generalizes to a support-ticket routing stream, where it improves accuracy by 18\% over reflection-free Thompson Sampling and by 51\% over the best prior baseline. Mechanism studies further show that the gains are causal: reflection helps precisely when regimes demand different retrieval behavior, and is provably no-harm/no-gain when the best policy is stable. Code is available at \footnote{\url{https://github.com/WujiangXu/AEL}}.
\end{abstract}

\section{Introduction}
\label{sec:intro}

LLM agents increasingly operate in long-horizon settings such as coding \citep{jimenez2024swebench,code-harness}, web navigation \citep{zhou2024webarena}, recommendation~\citep{iagent} and sequential prediction \citep{yao2023react}.
Around the frozen model sits a \emph{harness} \citep{code-harness}: the planner \citep{huang2024planning}, tool set \citep{schick2024toolformer}, and memory system \citep{xu2025mem} that decide what the model sees and does.
Hand-written and then fixed, it leaves the agent solving each episode with little reliable improvement from prior experience.
We argue that the bottleneck is the harness, not the model: not storing more memory, but learning \emph{how to use} it---which retrieval policy to apply, when past episodes are misleading, and when the harness itself should change.

\begin{figure}[tb!]
    \centering
    \begin{subfigure}[b]{\columnwidth}
        \centering
        \includegraphics[width=0.75\linewidth]{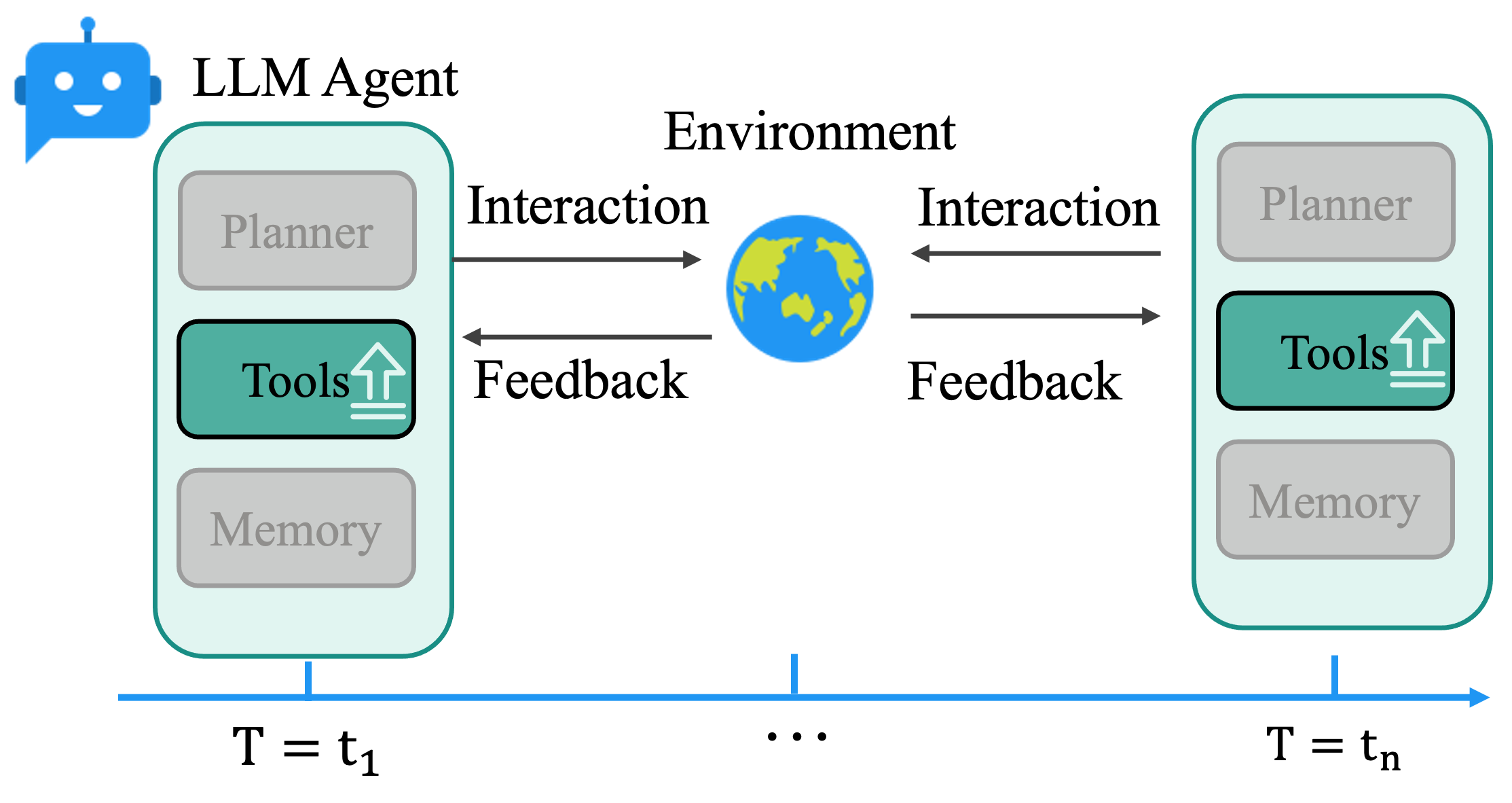}
        \caption{Previous methods~\citep{zhao2024expel,evotool2026}.}
        \label{fig:sub1}
    \end{subfigure}
    \vspace{1pt}
    \begin{subfigure}[b]{\columnwidth}
        \centering
        \includegraphics[width=0.85\linewidth]{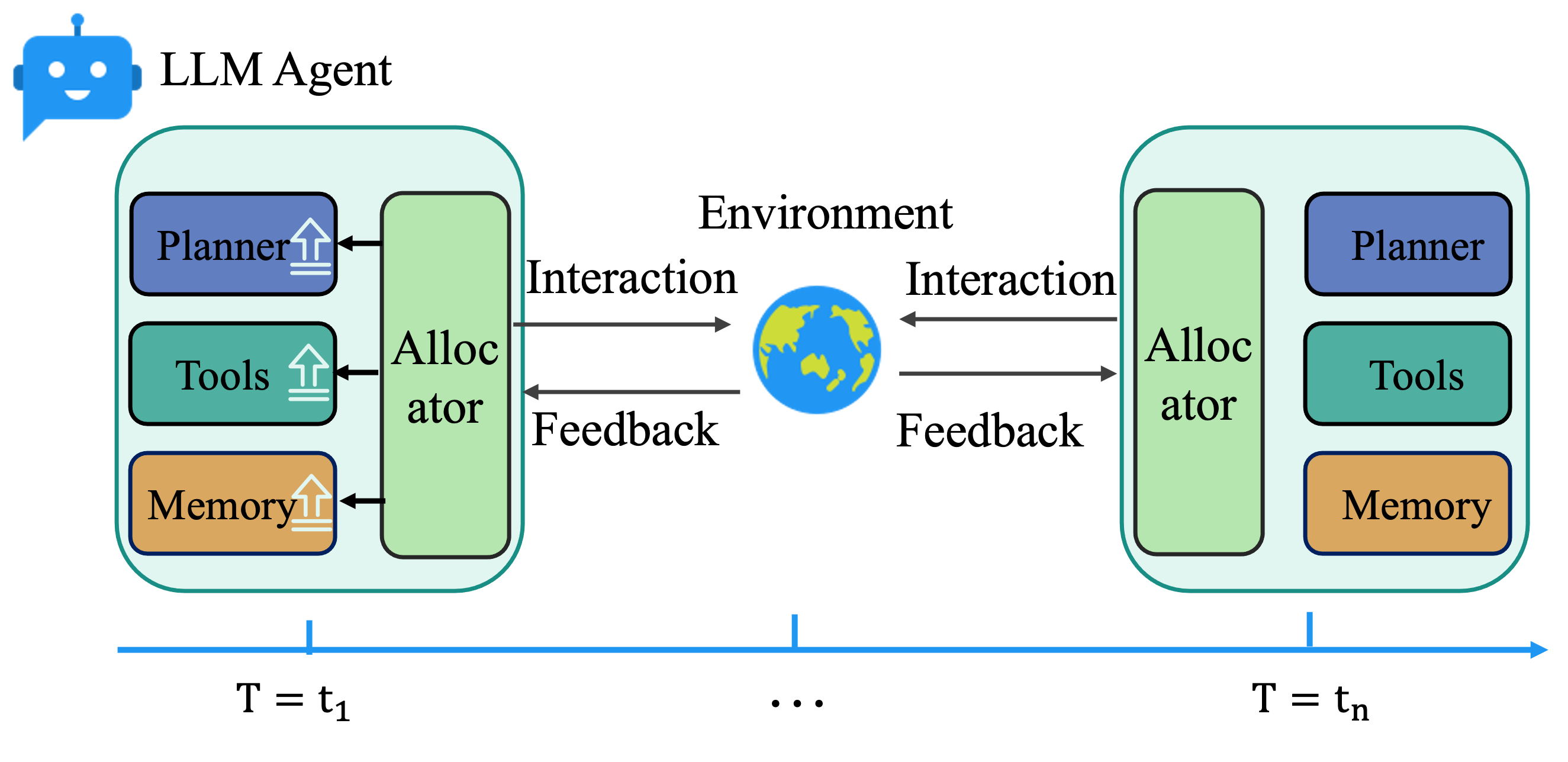}
        \caption{Our method.}
        \label{fig:sub2}
            \vspace{-8pt}
    \end{subfigure}
    \caption{Prior methods fix the harness to one retrieval/update pattern. \ael{} instead evolves the harness: it learns \emph{how} to retrieve experience with a fast memory-policy bandit and slow failure-diagnosis reflection.}
    \label{fig:main}
    \vspace{-15pt}
\end{figure}

Prior self-improving agents improve one representation at a time.
Reflexion \citep{shinn2023reflexion} accumulates critiques, ExpeL \citep{zhao2024expel} stores lessons, Meta-Reflexion \citep{metareflexion2025} distills rules, and EvoTool \citep{evotool2026} evolves tool policies.
These mechanisms are valuable, but they usually commit to a fixed retrieval or update pattern.
In dynamic sequential environments, this is brittle: the memory policy that helps early may overload the planner later, while a conservative policy may miss regime-specific evidence.

We introduce \emph{Agent Evolving Learning} (\ael{}), a two-timescale framework that evolves the harness rather than the model: the LLM stays frozen, and \ael{} learns the harness decision that matters most here, how to use memory.
At the fast timescale, Thompson Sampling \citep{chapelle2011thompson} chooses among memory-retrieval policies and updates each policy's posterior from scalar episode rewards.
At the slow timescale, LLM reflection aggregates trajectories, diagnoses failure modes, and injects a causal frame into the planner prompt; when the current policy pool underperforms, reflection may add a new retrieval policy as a bandit arm.
Evolving the other harness components, planner and tools, is supported but hurts in this short-horizon regime, so the default \ael{} keeps them fixed.
Our contributions are as follows.

\ding{182} \textbf{\textit{Framework.}} We propose \ael{}, a two-timescale framework that evolves an agent's harness (planner, tools, memory) with the LLM frozen. A fast Thompson Sampling bandit selects among memory-retrieval policies episode-by-episode, while slow LLM-driven reflection aggregates trajectories to diagnose failures and evolve the policy pool.

\ding{183} \textbf{\textit{Methodology.}} We introduce a \emph{diagnose-before-prescribe} reflection mechanism: rather than accumulating experience indiscriminately or evolving modules on a fixed schedule, the LLM first builds a causal explanation of why performance degraded, then generates a targeted architectural change. This couples qualitative failure diagnosis with quantitative bandit selection.

\ding{184} \textbf{\textit{Results.}} On the D-full portfolio benchmark, \ael{} achieves Sharpe $2.13$, outperforming ten baselines with the lowest variance among stochastic methods; on the support-ticket routing stream, \ael{} reaches $0.840$ accuracy, outperforming the strongest prior baseline by $+28.3$pp.


\section{Related Work}
\label{sec:related}

\subsection{LLM Agents}

ReAct \citep{yao2023react} interleaves chain-of-thought reasoning \citep{wei2022chain} with action execution, establishing the \emph{reason-then-act} paradigm.
Tree of Thoughts \citep{yao2024tree} expands the reasoning space via branching, while \citet{huang2024planning} identify planning as a key bottleneck.
Toolformer \citep{schick2024toolformer} demonstrates autonomous API invocation.
For memory, MemGPT \citep{packer2023memgpt} introduces OS-inspired paging between fast and slow storage, and A-Mem \citep{xu2025mem} proposes self-organized agentic memory.
These three modules (planner, tools, memory) form the agent harness, which is typically fixed once deployed. \ael{}'s default configuration learns over the memory subsystem (which retrieval policy to use) while keeping the planner and tools fixed; the framework also supports contextual bandits over planners and per-tool Thompson Sampling as ablation knobs.

\subsection{Evolving Learning of LLM Agents}

Several methods enable agents to improve, each evolving a subset of modules.
Reflexion \citep{shinn2023reflexion} accumulates self-critiques in the context window without structured retrieval.
Voyager \citep{wang2023voyager} builds a skill library through curriculum-driven exploration but does not adapt tools or planning.
ExpeL \citep{zhao2024expel} extracts lessons via keyword matching. Meta-Reflexion \citep{metareflexion2025} distills reflections into rules (the closest analogue to \ael{}'s procedural memory) but neither evolves tools or planners.
AutoAgent \citep{autoagent2026} explores elastic memory with evolving cognition.
EvoTool \citep{evotool2026} evolves tool-use policies via blame-aware mutation but holds memory fixed; FactorMiner \citep{factorminer2026} combines skills with experience memory; Tool-Genesis \citep{toolgenesis2026} benchmarks tool creation.
None jointly evolve tools and memory with reflection-driven self-diagnosis or study credit assignment in multi-module evolution.

\subsection{Bandits and Credit Assignment}

Contextual bandits \citep{li2010linucb,auer2002finite} and Thompson Sampling \citep{chapelle2011thompson,agrawal2012thompson} provide efficient online learning with natural exploration.
For multi-party credit assignment, Shapley values \citep{shapley1953value,lundberg2017shap,ghorbani2019data} offer a principled cooperative-game framework but largely treat modules as black boxes.
\ael{} systematically evaluates uniform, factored counterfactual, and LLM-driven credit assignment, finding that simpler methods often outperform in high-noise domains and further highlighting credit assignment as an important open challenge.

\begin{table*}[t]
\caption{Feature comparison with prior methods. \ael{} is the only method combining multi-tier memory, learned retrieval, and causal diagnosis across two timescales.}
\vspace{-10pt}
\label{tab:comparison}
\centering
\small
\renewcommand{\arraystretch}{1.15}
\setlength{\tabcolsep}{1pt}
\begin{tabular}{@{}l c c c c c@{}}
\toprule
Method & Memory & Retrieval & Diagnosis & Dual & Policy \\
       &        & Learning  &           & Scale & Evol. \\
\midrule
Reflexion~\citep{shinn2023reflexion}      & Context  & -- & --    & -- & -- \\
Voyager~\citep{wang2023voyager}         & Skills   & -- & --    & -- & -- \\
ExpeL~\citep{zhao2024expel}           & Lessons  & -- & --    & -- & -- \\
Meta-Reflexion~\citep{metareflexion2025}  & Rules    & -- & --    & -- & -- \\
EvoTool~\citep{evotool2026}         & --       & -- & blame & -- & tools \\
FactorMiner~\citep{factorminer2026}     & Experience     & -- & --    & -- & -- \\
\midrule
\rowcolor[gray]{0.94}
\ael{} (ours)   & \textbf{3-tier memory} & \textbf{\checkmark} & \textbf{\checkmark} & \textbf{\checkmark} & \textbf{\checkmark} \\
\bottomrule
\end{tabular}
\vspace{-10pt}
\end{table*}

\section{The \ael{} Framework}
\label{sec:method}

\subsection{Overview and Design Rationale}

\begin{figure*}[tb!]
\centering
\includegraphics[width=0.85\textwidth]{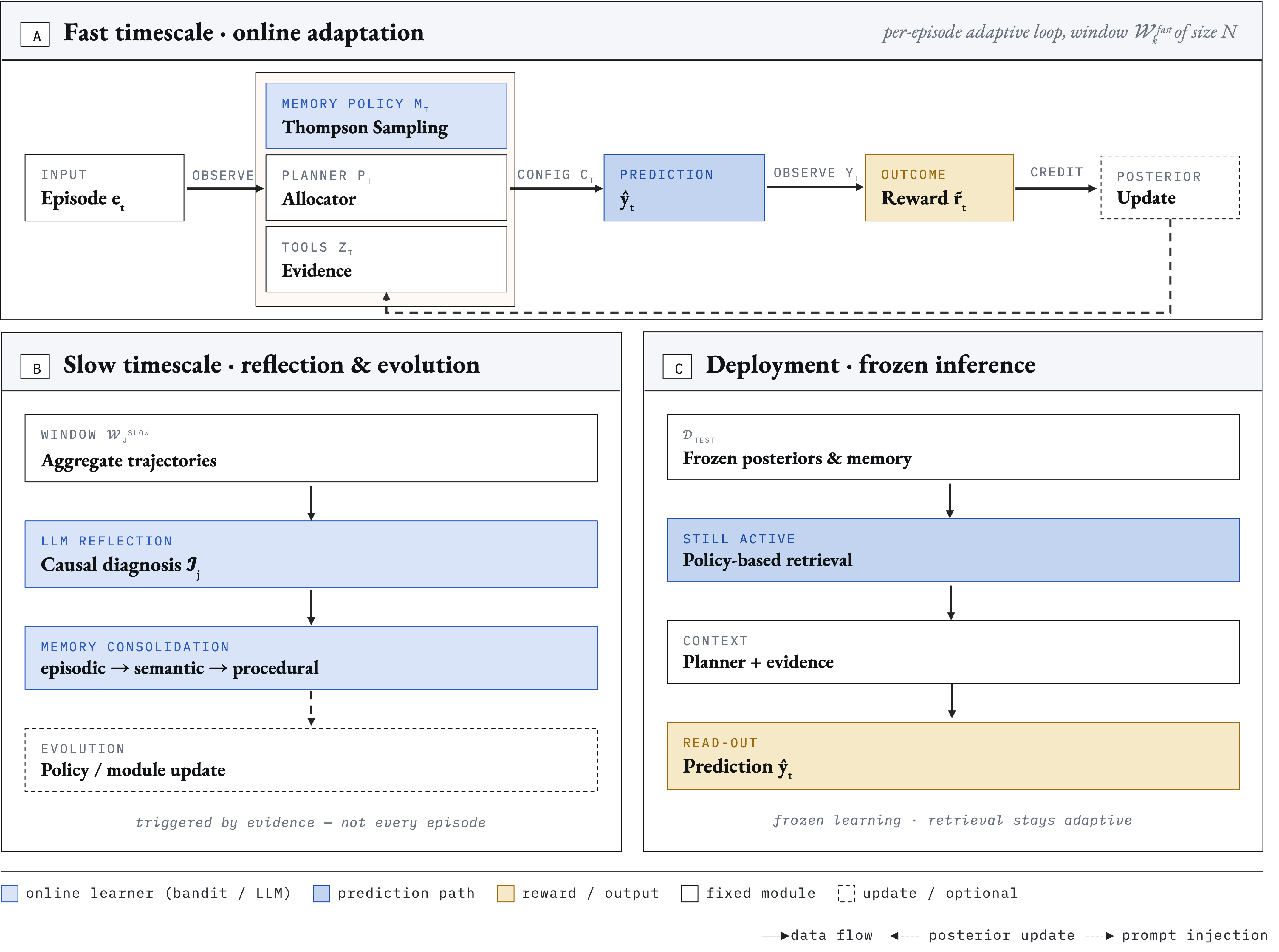}
\caption{AEL framework overview. At the fast timescale (A), a Thompson Sampling bandit selects memory retrieval policies episode-by-episode, feeding planner-guided predictions and bandit updates. At the slow timescale (B), an LLM reflects on aggregated trajectories to consolidate memory and evolve modules; during deployment (C), all weights are frozen while policy-based retrieval remains active.}
\label{fig:architecture}
\vspace{-10pt}
\end{figure*}

Let $\{e_t\}_{t=1}^T$ denote the episode stream.
\ael{} operates on two episode-based timescales: fast windows $\mathcal{W}^{\text{fast}}_k$ of size $N$ update selection preferences, while slow windows $\mathcal{W}^{\text{slow}}_j$ of size $M \gg N$ aggregate evidence for reflection and memory consolidation.
At episode $t$, the agent selects a harness configuration $c_t = (p_t, z_t, m_t)$ specifying planner, tool set, and memory retrieval policy, then observes outcome score $s_t \in [-1,1]$.
In the default \ael{} configuration, the planner and tools are fixed; a Thompson Sampling bandit selects among memory retrieval policies, learning which way of accessing experience suits the current stage.
At the slow timescale, LLM-driven reflection diagnoses failure patterns and injects causal insights into the decision prompt.
Uniform credit converts outcomes into bandit rewards.
\autoref{fig:architecture} shows the pipeline; episodes are split chronologically into train/val/test, with all learning frozen at test time.

\paragraph{Which parts of the harness are learned.}
The default \ael{} configuration learns three quantities online:
(i)~a Beta posterior per memory-retrieval policy via Thompson Sampling,
(ii)~a reflection insight regenerated each slow window and injected into the planner prompt, and
(iii)~procedural rules promoted from recurring semantic patterns.
The planner identity, the tool set (12 tools), and all tool implementations remain fixed.
The framework additionally supports planner selection (LinUCB contextual bandit), per-tool Thompson Sampling, cold-start initialization (with uninformative priors $\alpha_0{=}\beta_0{=}1$), and skill extraction as optional extensions; we keep all of these off by default because the ablation (\autoref{tab:ablation}) shows they degrade performance in this short-horizon regime.
Credit assignment uses uniform rewards; factored counterfactual (\fcc{}) and LLM-driven (\llmfcc{}) alternatives are also evaluated in \autoref{tab:ablation} and found harmful.

\subsection{Memory-Policy Selection and Extended Module Bandits}

The central online learning mechanism in \ael{} is memory-policy selection via Thompson Sampling \citep{chapelle2011thompson}.
Each policy $m$ in a pool of five initial strategies maintains a Beta posterior $\mathrm{Beta}(\alpha_m, \beta_m)$; at each episode the agent samples $\tilde{\mu}_m \sim \mathrm{Beta}(\alpha_m, \beta_m)$ and selects the policy with the highest sample.
After observing the episode outcome, the posterior is updated: $\alpha_m \mathrel{+}= \tilde{r}_t$, $\beta_m \mathrel{+}= (1 - \tilde{r}_t)$.
The five initial policies range from no retrieval to aggressive multi-tier recall (\autoref{app:planner-memory}); new policies can be added by the evolution mechanism described in \autoref{sec:reflection}.
This learns \emph{how} to access experience (compressed summaries, full records, or no retrieval), adapting the strategy as the memory store matures. The framework additionally supports two further bandit levels, evaluated in the ablation (\autoref{tab:ablation}):

\noindent \textbf{Planner selection (LinUCB \citep{li2010linucb}).}
A contextual bandit whose context vector $\phi_t \in \mathbb{R}^7$ encodes sector, 30-day volatility, log market cap, data richness, momentum, options availability, and analyst coverage.
It selects $\pi_t = \arg\max_\pi \bigl(\phi_t^\top \hat{\theta}_\pi + \alpha\sqrt{\phi_t^\top \mathbf{A}_\pi^{-1}\phi_t}\bigr)$, balancing exploitation of the best-performing planner with exploration of uncertain alternatives (full update equations in \autoref{app:bandit-details}).

\noindent \textbf{Per-tool selection (Thompson Sampling).}
Each tool $a$ maintains a Beta posterior $\mathrm{Beta}(\alpha_a, \beta_a)$ updated from directional hit/miss statistics against realized outcomes; the top-$K$ tools by sampled value are selected, with $K$ shrinking as learning progresses (details in \autoref{app:bandit-details}).

\subsection{Three-Tier Evolving Memory}

Raw episode logs are too noisy and voluminous to retrieve directly. The agent needs to progressively distill experience from specific episodes into general patterns and then into actionable rules.
\ael{} achieves this through three memory tiers with automatic promotion.

\noindent \textbf{Episodic memory} records each episode's raw outcome: which tools were used, what signals they produced, whether each signal was correct, and the overall prediction quality.
This provides the ground-truth training data for distillation.

\noindent \textbf{Semantic memory} aggregates episodic records into cross-episode patterns, distilled periodically (every 10 episodes).
These patterns capture regularities invisible in any single episode, such as ``momentum indicators are reliable for trending stocks but misleading during reversals.''

\noindent \textbf{Procedural memory} promotes high-confidence semantic patterns into executable rules injected directly into planner prompts.
These rules represent the agent's most trusted knowledge, influencing behavior without requiring retrieval at query time.

Memory use involves two distinct decisions.
First, the Thompson Sampling bandit selects a \emph{retrieval policy} that specifies which tiers are visible, how many entries to retrieve, and how they are formatted (e.g., compressed summaries vs.\ full records vs.\ no retrieval at all).
Second, \emph{within} the selected policy, entries are ranked by a composite relevance score:
\begin{equation*}
\resizebox{\columnwidth}{!}{$
  r(q, e) = \underbrace{f_{\text{match}}(q, e)}_{\text{ticker, sector, tool}} \!\times\! \underbrace{(0.5 + 0.5 q_e)}_{\text{quality}} \!\times\! \underbrace{(0.3 + 0.7 e^{-0.01 \Delta})}_{\text{recency}} \!\times\! \underbrace{b_{\tau}}_{\text{tier boost}}
$}
\end{equation*}
where $f_{\text{match}}$ sums feature-match bonuses, $q_e$ is the entry's quality score, $\Delta$ is the number of episodes since the entry was written, and $b_\tau \in \{1.0, 1.2, 1.5\}$ for episodic, semantic, and procedural tiers respectively.
The top-$k$ entries (default $k{=}5$) above a quality threshold are returned.
This two-layer design means the bandit learns \emph{how} to access experience, while the scoring function determines \emph{which} entries surface under that access strategy.

\subsection{LLM-Driven Reflection and Code Evolution}
\label{sec:reflection}

Selection within a fixed module pool will eventually plateau if the pool itself is inadequate.
The reflection system addresses this by letting the LLM diagnose \emph{why} performance has degraded and then generate new modules to address the diagnosed problem.

\noindent \textbf{Cold-start initialization.}
Before any episodes, the LLM reads tool and planner descriptions and generates informed priors for the bandit system, reducing the exploration cost of uniform initialization.

\noindent \textbf{Slow-window reflection.}
After each slow window, the LLM receives four inputs: (i)~per-ticker episode summaries (planner used, score, directional accuracy), (ii)~per-tool accuracy statistics aggregated over the window, (iii)~market side information (sector returns, volatility regime, cross-correlation) computed from cached price data but \emph{not} given to the predictor, and (iv)~the last three reflections for temporal continuity.
It produces a structured output containing a \emph{causal insight} (what market conditions caused today's outcomes and why specific signals were reliable or misleading), a regime label, and a confidence score.
This insight is injected directly into the allocator's prompt at the next episode, giving the predictor an interpretive frame for its evidence without polluting the memory store with noisy LLM-generated narratives.
Module selection remains with the bandit; reflection diagnoses conditions, not configurations.

\noindent \textbf{Code evolution} is triggered by diagnosed structural failures, not fixed schedules.
When a planner's failure streak exceeds a threshold (default 3 consecutive slow windows), the LLM generates a new Python planner class tailored to the diagnosed failure mode; the new class must pass syntax validation and a smoke test before entering the pool.
Memory policy evolution fires periodically (every 5 slow windows, starting after window 10) only when the average Thompson Sampling reward across all policies falls below 0.4, indicating that the entire policy pool is underperforming.
The LLM then designs a new retrieval policy specifying tier visibility, retrieval depth, and formatting strategy, which is added as a new bandit arm.
Note that in the main \ael{} configuration, planner evolution is disabled; it is active only in the full EAEL variant.
The design principle is \emph{diagnose before prescribe}: the LLM must first build an explanation of why performance degraded before generating an architectural change, making evolution targeted rather than random.

\subsection{A Running Case: One Failure, One Reflection, One Policy Shift}
\label{sec:running-case}

On seed~42, a bear-regime loss causes the memory bandit to down-weight an \texttt{aggressive\_learner} retrieval policy.
After the slow window, reflection diagnoses that bull-regime memories are misleading under elevated volatility and later proposes \texttt{RegimeFilteredRetrieval}, a small validated retrieval policy that filters memories by regime metadata.
The new arm enters Thompson Sampling with uninformative priors; within the next slow window the mean reward rises from $0.19$ to $0.38$.
This trace illustrates the intended division of labor: the bandit consumes scalar rewards, while reflection consumes aggregated trajectories and side information to decide whether the policy pool itself is missing an arm.
The full trace and generated code are in \autoref{app:running-case-full} and \autoref{app:code-examples}.

\subsection{Learning Signal}
\label{sec:credit}

The training signal combines two complementary mechanisms at different timescales.
At the fast timescale, each episode produces a scalar outcome $s_t \in [-1,1]$ that is converted into a \textbf{uniform reward} $\tilde{r}_t = \mathrm{clip}((s_t + 1)/2, 0, 1)$ and used directly to update the memory-policy bandit posterior (and, in the full EAEL variant, tool and planner bandits as well).
Uniform credit avoids compounding misattribution in our high-noise domain. We also evaluate factored counterfactual credit (\fcc{}, combining structural, counterfactual, and Shapley attribution) and LLM-driven credit (\llmfcc{}), but both degrade performance (\autoref{tab:ablation}; full formulations in \autoref{app:credit-details}).
At the slow timescale, reflection consumes aggregated trajectories to decide \emph{when structural evolution is needed}, providing a qualitative diagnostic channel that complements the quantitative bandit signal.

\begin{algorithm}[t]
\caption{The training process of \ael{}}\label{alg:train}
\small
\begin{algorithmic}[1]
\Require Episode stream $\{e_t\}_{t=1}^T$; memory-policy pool $\mathcal{M}{=}\{m_1,\ldots,m_K\}$; priors $\alpha_k{=}\beta_k{=}1$; slow-window size $M$; warm-up $W$; evolution threshold $\bar{r}_{\min}$
\Ensure Posteriors $\{(\alpha_k,\beta_k)\}$, memory $\mathcal{D}$, pool $\mathcal{M}$
\State Partition $\{e_t\}$ into $\mathcal{E}_{\mathrm{train}},\mathcal{E}_{\mathrm{val}},\mathcal{E}_{\mathrm{test}}$ chronologically
\For{each slow window $\mathcal{W}_j \subset \mathcal{E}_{\mathrm{train}}$}
    \For{each episode $e_t \in \mathcal{W}_j$}
        \State Sample $\tilde{\mu}_k \sim \mathrm{Beta}(\alpha_k,\beta_k)$ for each $m_k \in \mathcal{M}$
        \State $m_t \gets \arg\max_k \tilde{\mu}_k$ \Comment{Thompson Sampling}
        \State $\mathbf{x}_t \gets \textsc{Retrieve}(\mathcal{D}, m_t, e_t)$ \Comment{Policy-guided recall}
        \State $\hat{y}_t \gets \textsc{Plan}(e_t, \mathbf{x}_t, \mathcal{Z})$ \Comment{Predict with tools $\mathcal{Z}$}
        \State Observe $y_t$; compute $s_t \in [-1,1]$
        \State $\tilde{r}_t \gets \mathrm{clip}\bigl((s_t{+}1)/2,\;0,\;1\bigr)$ \Comment{Uniform credit}
        \State $\alpha_{m_t} \mathrel{+}= \tilde{r}_t$;\; $\beta_{m_t} \mathrel{+}= 1{-}\tilde{r}_t$
        \State Write $(e_t,\hat{y}_t,y_t,s_t)$ to episodic tier of $\mathcal{D}$
    \EndFor
    \State $\mathcal{I}_j \gets \textsc{Reflect}(\mathcal{W}_j, \mathcal{D})$ \Comment{LLM causal diagnosis}
    \State Inject $\mathcal{I}_j$ into planner prompt for $\mathcal{W}_{j+1}$
    \State Distill episodic $\to$ semantic $\to$ procedural tiers of $\mathcal{D}$
    \If{$\bar{r}_{\mathcal{M}} < \bar{r}_{\min}$ \textbf{and} $j > j_{\min}$}
        \State $m_{\mathrm{new}} \gets \textsc{EvolvePolicy}(\mathcal{I}_j, \mathcal{D})$; $\mathcal{M} \gets \mathcal{M} \cup \{m_{\mathrm{new}}\}$
    \EndIf
\EndFor
\State Select best posteriors on $\mathcal{E}_{\mathrm{val}}$; freeze for $\mathcal{E}_{\mathrm{test}}$
\end{algorithmic}
\end{algorithm}

\section{Experiments}
\label{sec:experiments}

We design experiments to answer four research questions.
\textbf{Q1}:~Does \ael{} outperform prior self-improving agent methods?
\textbf{Q2}:~Which components contribute, and does adding complexity help or hurt?
\textbf{Q3}:~Does the two-timescale mechanism generalize beyond finance?
\textbf{Q4}:~Are the improvements statistically significant?
Additional mechanism analyses (bandit alternatives, reflection faithfulness, procedural-memory dynamics) are in the appendix.

\subsection{Experiments Setup}

\noindent \textbf{Datasets.} We evaluate on two benchmarks.
\textbf{(1)}~\emph{D-full portfolio}: 10 sector-diverse tickers spanning 7 GICS sectors at 1-hour resolution (208 episodes: 140 train / 40 val / 28 test). Training covers diverse market regimes (bull, bear, flat); the test set contains a bear-to-bull transition (full details in \autoref{app:dataset}).
\textbf{(2)}~\emph{Support-ticket routing}: an LLM-generated stream with 8 resolution routes, produced via Gemini 3.1 Flash-Lite. The stream spans four regimes: explicit logs, shorthand drift, imbalanced incidents, and a late route-aware regression. The LLM-agent evaluation uses 160 stratified episodes per seed (40 per regime) across 5 seeds (details in \autoref{sec:crossdomain-cliapi}).

\noindent \textbf{Baselines.} We compare:
(i)~four \textbf{non-LLM baselines} (equal-weight, momentum-weighted, min-variance, inverse-momentum);
(ii)~five \textbf{prior self-improving methods}: Reflexion~\citep{shinn2023reflexion}, ExpeL~\citep{zhao2024expel}, FactorMiner~\citep{factorminer2026}, Meta-Reflexion~\citep{metareflexion2025}, and EvoTool~\citep{evotool2026} (adaptation details in \autoref{app:baselines});
(iii)~\textbf{HyperAgent}~\citep{zhang2026hyperagents}, a recursive self-modification framework (\autoref{sec:casestudy}); and
(iv)~an \textbf{\ael{} incremental build} (Stateless $\to$ +Tools $\to$ +Memory $\to$ \ael{}).
All methods share the same 12 tools (\autoref{app:tools}), backbone LLM (Claude Haiku 4.5), and data split.

\noindent \textbf{Evaluation \& Protocol.}
\label{sec:metrics}
\label{sec:implementation}
We report four frozen test-phase metrics: Sharpe, Sortino, and Calmar ratios (risk-adjusted return from complementary perspectives) and maximum drawdown (MaxDD\%); formal definitions are in \autoref{app:metrics}.
The matched incremental build uses seeds 42, 123, 456; headline methods extend to 5 seeds.
During test, all learning is disabled (frozen bandits, read-only memory, no evolution).
More details are in \autoref{app:protocol}.

\subsection{Main Results (Q1)}
\label{sec:results}

\begin{table*}[t]
\caption{Main results on two benchmarks ($N{=}5$ seeds for all stochastic methods). Bold = best per column.}
\label{tab:main_results}
\centering
\small

\begin{subtable}{\textwidth}
\vspace{-10pt}
\caption{D-full portfolio benchmark. Non-LLM baselines: EqW (equal-weight), Mom (momentum-weighted), MinV (min-variance), InvM (inverse-momentum). Metric definitions in \autoref{app:metrics}.}
\label{tab:main}
\centering
\setlength{\tabcolsep}{5pt}
\renewcommand{\arraystretch}{1.08}
\begin{tabular}{@{}llcccc@{}}
\toprule
Category & Method & Sharpe $\uparrow$ & Sortino $\uparrow$ & Calmar $\uparrow$ & MaxDD\% $\uparrow$ \\
\midrule
\multirow{4}{*}{Non-LLM} & EqW & 0.70 & 1.32 & 3.05 & $-$1.47 \\
 & Mom & 1.44 & 2.73 & 6.89 & $-$1.64 \\
 & MinV & $-$0.61 & $-$0.82 & $-$2.06 & $-$1.59 \\
 & InvM & $-$0.20 & $-$0.30 & $-$0.66 & $-$1.73 \\
\midrule
\multirow{6}{*}{Prior} & Reflexion~\citep{shinn2023reflexion} & $-$0.59$\pm$1.33 & $-$0.57$\pm$1.75 & $-$1.64$\pm$5.92 & $-$2.30$\pm$0.36 \\
 & ExpeL~\citep{zhao2024expel} & 0.76$\pm$1.93 & 1.68$\pm$3.93 & 3.97$\pm$9.79 & $-$1.76$\pm$0.16 \\
 & FactorMiner~\citep{factorminer2026} & 0.85$\pm$1.15 & 1.55$\pm$2.14 & 3.61$\pm$4.91 & $-$1.93$\pm$0.28 \\
 & Meta-Reflection~\citep{metareflexion2025} & 0.20$\pm$1.16 & 0.37$\pm$1.65 & 1.66$\pm$6.01 & $-$2.22$\pm$0.23 \\
 & EvoTool~\citep{evotool2026} & 1.37$\pm$1.74 & 2.73$\pm$3.58 & 6.28$\pm$7.70 & \textbf{$-$1.46$\pm$0.15} \\
 & HyperAgent~\citep{zhang2026hyperagents} & 0.72$\pm$0.38 & 1.00$\pm$0.53 & 3.90$\pm$2.37 & $-$2.49$\pm$0.15 \\
\midrule
\multirow{3}{*}{\ael{} Variants} & Stateless & 1.35$\pm$1.03 & 2.51$\pm$1.93 & 6.52$\pm$4.98 & $-$1.59$\pm$0.26 \\
 & +Tools & 1.29$\pm$1.19 & 2.51$\pm$2.23 & 6.07$\pm$5.52 & $-$1.63$\pm$0.26 \\
 & +Memory & 1.68$\pm$0.96 & 3.29$\pm$1.83 & 8.50$\pm$5.09 & $-$1.53$\pm$0.11 \\
\midrule
\rowcolor[gray]{0.94}
\textbf{Ours} & \textbf{\ael{}} & \textbf{2.13$\pm$0.47} & \textbf{4.08$\pm$1.11} & \textbf{10.40$\pm$2.75} & $-$1.53$\pm$0.06 \\
\bottomrule
\end{tabular}
\end{subtable}

\vspace{6pt}

\begin{subtable}{\textwidth}
\caption{Support-ticket routing (Gemini 3.1 Flash-Lite, 160 stratified episodes per seed). R0--R3 are explicit logs, shorthand drift, imbalanced incidents, and route-aware regression. Baseline adaptations in \autoref{app:cliapi-details}.}
\label{tab:cliapi_llm_text}
\centering
\setlength{\tabcolsep}{3pt}
\begin{tabular}{@{}llccccc@{}}
\toprule
Category & Method & Overall & R0 & R1 & R2 & R3 \\
\midrule
No memory & Stateless & 0.512 $\pm$ 0.010 & 0.970 & 0.235 & 0.220 & 0.625  \\
Fixed policy & Class-balanced & 0.565 $\pm$ 0.036 & 0.965 & 0.290 & 0.465 & 0.540  \\
\midrule
\multirow{5}{*}{Prior} & Reflexion~\citep{shinn2023reflexion} & 0.490 $\pm$ 0.045 & 0.925 & 0.230 & 0.235 & 0.570  \\
 & ExpeL~\citep{zhao2024expel} & 0.539 $\pm$ 0.018 & 0.960 & 0.245 & 0.245 & 0.705  \\
 & Meta-Reflection~\citep{metareflexion2025} & 0.497 $\pm$ 0.031 & 0.980 & 0.245 & 0.295 & 0.470  \\
 & EvoTool~\citep{evotool2026} & 0.540 $\pm$ 0.046 & 0.980 & 0.370 & 0.435 & 0.375  \\
 & FactorMiner~\citep{factorminer2026} & 0.557 $\pm$ 0.037 & 0.960 & 0.290 & 0.455 & 0.525  \\
\midrule
Ablation & TS no reflection & 0.711 $\pm$ 0.039 & \textbf{0.985} & 0.590 & 0.585 & 0.685  \\
\midrule
\rowcolor[gray]{0.94}
\textbf{Ours} & \textbf{\ael{}} & \textbf{0.840 $\pm$ 0.031} & \textbf{0.985} & \textbf{0.775} & \textbf{0.715} & \textbf{0.885} \\
\bottomrule
\end{tabular}
\end{subtable}

\vspace{-10pt}
\end{table*}

\autoref{tab:main} reports frozen-test metrics on the D-full benchmark across five random seeds.
We focus on the Sharpe ratio as the primary metric because it directly reflects the training signal (episode-level return normalized by volatility) and captures the risk-return tradeoff that is central to sequential portfolio allocation.
Sortino and Calmar ratios provide complementary views on downside risk and worst-case drawdown.
\ael{} achieves the highest Sharpe (2.13), Sortino (4.08), and Calmar (10.40), outperforming all 10 baselines with the lowest variance among all stochastic methods.

\noindent \textbf{Comparison to prior methods.}
Among prior methods, EvoTool is the strongest on mean Sharpe (1.37) but with the highest variance, indicating that its evolutionary tool policy is sensitive to initialization.
ExpeL (0.76) and Reflexion ($-$0.59) both exhibit extreme seed dependence: individual outlier seeds inflate their means (detailed in \autoref{app:results-detail}).
HyperAgent (0.72) achieves the lowest variance among prior methods, but its recursive code generation fails to escape the initial equal-weight template (\autoref{app:casestudy-detail}).
The deterministic momentum-weighted baseline (1.44) outperforms all prior LLM methods on Sharpe, underscoring that LLM-based learning must overcome the noise it introduces to beat simple heuristics.

\noindent \textbf{Why \ael{} improves.}
The incremental build isolates the source of improvement: memory enables cross-episode learning (+24\%), and reflection further enhances performance (+27\%) by diagnosing failure patterns and filtering memory quality.
Unlike prior methods that accumulate experience indiscriminately (Reflexion) or evolve only one module (EvoTool), \ael{} uses reflection to provide the agent with an \emph{interpretive frame} for its evidence, turning raw memory into actionable diagnostic insights.
The low variance further suggests that this diagnostic mechanism generalizes across random initializations rather than depending on a seed.

\noindent \textbf{Cross-domain validation on support-ticket routing.}
On the support-ticket routing benchmark (\autoref{tab:cliapi_llm_text}), each episode is a support ticket for an LLM API service with four regimes (explicit logs, shorthand drift, imbalanced incidents, and route-aware regression).
Using Gemini 3.1 Flash-Lite with zero errors or fallbacks, \ael{} reaches $0.840$ accuracy, outperforming TS without reflection ($0.711$) and the strongest prior baseline (FactorMiner, $0.557$).
The result is consistent with the finance story: reflection helps when it introduces a retrieval behavior unavailable in the initial pool, particularly in the late route-aware regime where R3 accuracy reaches $0.885$ versus $0.705$ for the strongest prior baseline.

\subsection{Component Synergy and Robustness (Q2)}

\begin{figure}[t]
\centering
\begin{subfigure}[t]{0.4\linewidth}
    \includegraphics[width=\linewidth]{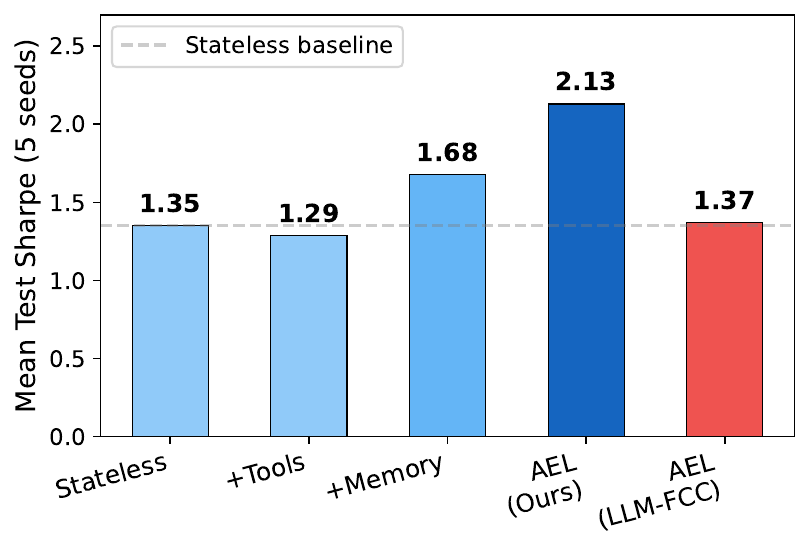}
    \caption{Incremental component build.}
    \label{fig:synergy}
\end{subfigure}
\hfill
\begin{subfigure}[t]{0.58\linewidth}
    \includegraphics[width=\linewidth]{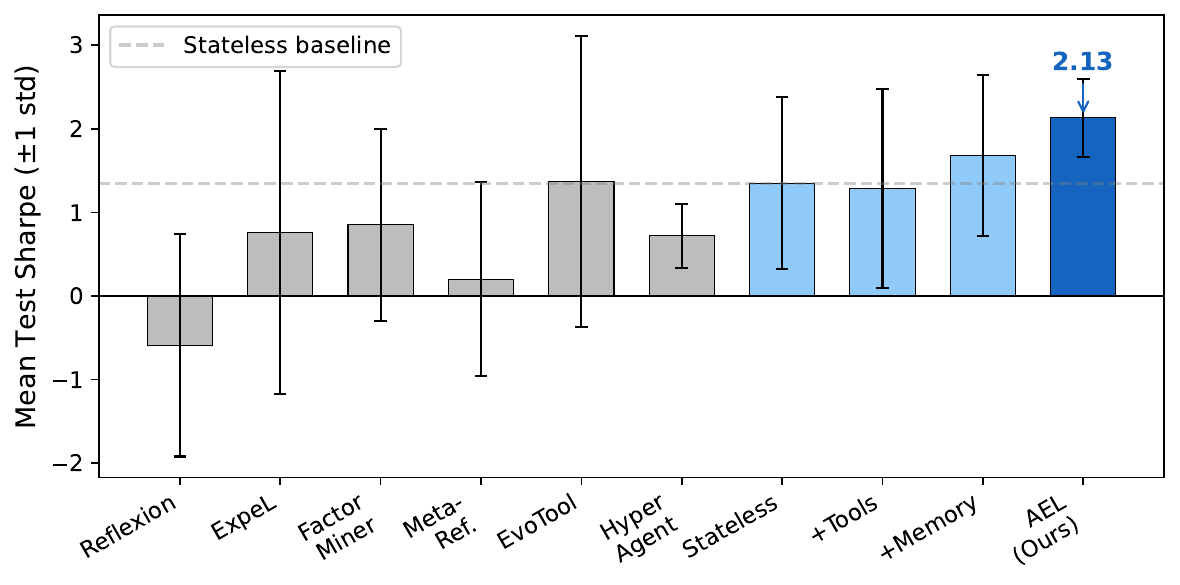}
    \caption{Mean-focused robustness summary.}
    \label{fig:seeds}
\end{subfigure}
\vspace{-7pt}
\caption{\textbf{(a) Incremental build (5 seeds):} Stateless (1.35) $\to$ +Memory (1.68) $\to$ \ael{} (\textbf{2.13}); adding LLM-FCC credit degrades to 1.37. \textbf{(b) Robustness:} Mean test Sharpe $\pm$1 std. \ael{} has the highest mean (2.13) with the lowest variance among LLM methods.}
\label{fig:synergy_and_seeds}
\vspace{-12pt}
\end{figure}

\autoref{fig:synergy} reveals two findings.
First, \emph{memory and reflection form a synergistic pair}: Stateless (1.35) $\to$ +Memory (1.68) $\to$ \ael{} (\textbf{2.13}), with reflection providing a 27\% jump over memory alone.
The disproportionate gain from reflection confirms that the bottleneck in evolving an agent harness is \emph{how to use} experience, not simply accumulating it.
Second, adding LLM-FCC credit to the full AEL system \emph{degrades} performance (1.37), showing that sophisticated credit assignment introduces more noise than signal in this high-noise domain.
\autoref{fig:seeds} confirms that \ael{} achieves the highest mean Sharpe with the lowest variance among LLM methods exceeding mean Sharpe $1.0$, while EvoTool, +Memory, and Stateless all exhibit substantially higher seed-level variability. HyperAgent attains a lower std but at a markedly lower mean (0.72), reflecting its collapse to an equal-weight prior rather than adaptive selection.

\subsection{Ablation and Credit Analysis (Q2)}
\label{sec:credit-analysis}

\begin{figure}[tb!]
\centering
\includegraphics[width=0.9\linewidth]{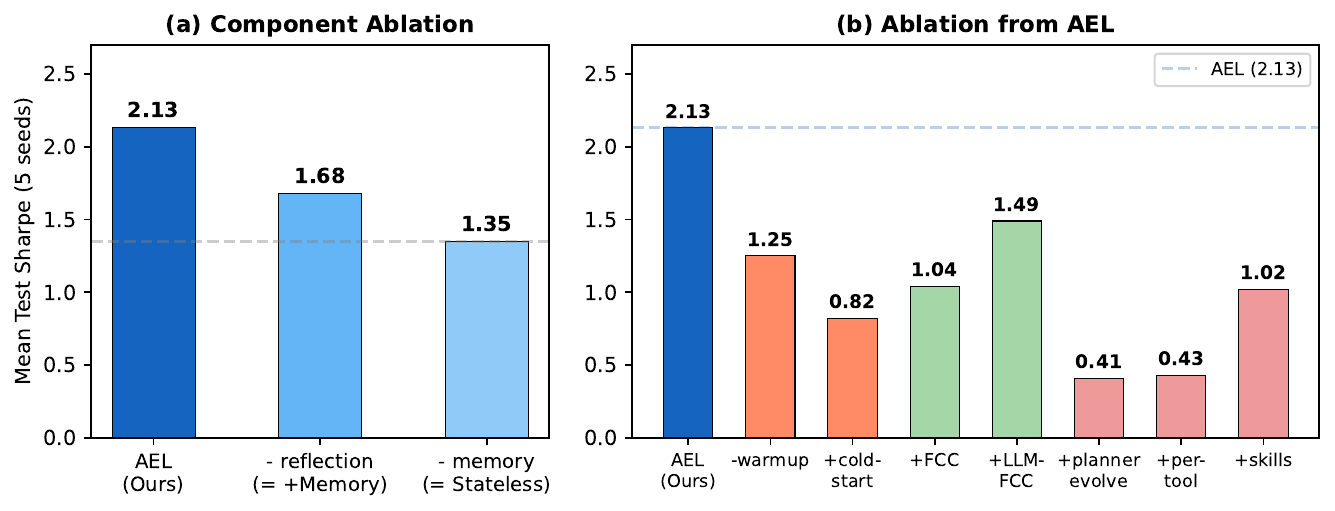}
\vspace{-10pt}
\caption{The ablation study.}
\label{fig:ablation}
\vspace{-10pt}
\end{figure}

\autoref{tab:ablation} and \autoref{fig:ablation} reveal a striking finding: \emph{every modification to \ael{} degrades performance}.
Removing warm-up ($\Delta$$-$0.88) is most damaging: without the 15-episode warm-up, the memory bandit receives noisy early rewards that corrupt Beta posteriors irreversibly, causing premature convergence to a suboptimal retrieval policy.
Reflection contributes $\Delta$$-$0.45 via failure diagnosis and quality filtering; though modest in magnitude, the +Memory baseline (1.68) is already strong, so reflection's value lies in interpretive quality rather than raw gain.
Adding complexity consistently hurts: planner evolution ($\Delta$$-$1.72) and per-tool selection ($\Delta$$-$1.70) suffer from data starvation (208 episodes split across 6 planners or 120 tool-ticker arms), cold-start ($\Delta$$-$1.31) introduces distributional bias, and skill extraction ($\Delta$$-$1.11) overfits to training regimes.
For credit assignment, both \fcc{} ($\Delta$$-$1.09) and \llmfcc{} ($\Delta$$-$0.64) underperform uniform credit, highlighting credit assignment as an open problem in high-noise, short-horizon settings.

\begin{table}[tb!]
\vspace{-7pt}
\caption{Ablation from \ael{} (5 seeds). Each row modifies one component. Every change degrades performance.}
\label{tab:ablation}
\centering
\small
\renewcommand{\arraystretch}{1.10}
\setlength{\tabcolsep}{5pt}
\begin{tabular}{@{}lccc@{}}
\toprule
Configuration & Sharpe & Sortino & $\Delta$Sharpe \\
\midrule
\rowcolor[gray]{0.94}
\ael{} (Ours) & \textbf{2.13} & \textbf{4.08} & \textcolor{gray}{---} \\
\midrule
\multicolumn{4}{@{}l}{\textcolor{gray}{\textit{Remove component}}} \\
\hspace{0.35em}$-$ warm-up & 1.25 & 2.25 & \textcolor{red!70!black}{$-$0.88} \\
\hspace{0.35em}$-$ reflection & 1.68 & 3.29 & \textcolor{red!70!black}{$-$0.45} \\
\midrule
\multicolumn{4}{@{}l}{\textcolor{gray}{\textit{Add complexity}}} \\
\hspace{0.35em}$+$ cold-start init & 0.82 & 1.41 & \textcolor{red!70!black}{$-$1.31} \\
\hspace{0.35em}$+$ planner evol. & 0.41 & 1.01 & \textcolor{red!70!black}{$-$1.72} \\
\hspace{0.35em}$+$ per-tool sel. & 0.43 & 0.80 & \textcolor{red!70!black}{$-$1.70} \\
\hspace{0.35em}$+$ skill extract. & 1.02 & 1.78 & \textcolor{red!70!black}{$-$1.11} \\
\midrule
\multicolumn{4}{@{}l}{\textcolor{gray}{\textit{Change credit method}}} \\
\hspace{0.35em}$\to$ \fcc{} & 1.04 & 1.96 & \textcolor{red!70!black}{$-$1.09} \\
\hspace{0.35em}$\to$ \llmfcc{} & 1.49 & 2.84 & \textcolor{red!70!black}{$-$0.64} \\
\bottomrule
\end{tabular}
\vspace{-10pt}
\end{table}

\subsection{Cross-Domain Mechanism Transfer (Q3)}
\label{sec:crossdomain-cliapi}
\label{sec:crossdomain-synth}

To test whether the two-timescale mechanism is domain-agnostic rather than finance-specific, we construct a deterministic, LLM-free 10-arm Bernoulli bandit with four regimes (52 episodes each, 208 total).
A hidden 11th arm pays $0.70$ during regime 3 but only $0.30$ in regimes 0--2; it is not available at start and can only be added through reflection.
\begin{table}[t]
\caption{Synthetic cross-domain bandit (5 seeds). Reflection-driven arm injection yields $+9.3$pp in R3.}
\label{tab:synth_crossdomain}
\centering
\small
\renewcommand{\arraystretch}{1.05}
\setlength{\tabcolsep}{3pt}
\begin{tabular}{@{}lcccc@{}}
\toprule
Algorithm & R0 & R3 & Overall & Sharpe \\
\midrule
RR          & .364 & .352 & .385\tiny$\pm$.023 & .79 \\
UCB1        & .392 & .410 & .402\tiny$\pm$.026 & .82 \\
$\varepsilon$-gr & .396 & .331 & .364\tiny$\pm$.017 & .76 \\
TS          & .372 & .359 & .385\tiny$\pm$.021 & .79 \\
\rowcolor[gray]{0.94}
\ael{}      & .408 & \textbf{.452} & .400\tiny$\pm$.020 & \textbf{.82} \\
\midrule
Oracle      & .632 & .731 & .688\tiny$\pm$.011 & 1.49 \\
\bottomrule
\end{tabular}
\vspace{-3pt}
\end{table}
\autoref{tab:synth_crossdomain} shows that \ael{}-style TS+reflection reaches $0.400$ overall, comparable to UCB1 ($0.402$), but in regime 3 where the hidden arm fires, \ael{} reaches $\mathbf{0.452}$ versus pure TS at $0.359$ --- a $+9.3$pp lift.
The mechanism transfers directionally, though the gain is smaller than in finance because the synthetic setting lacks informative regime features that reflection can exploit.
A 20Newsgroups boundary case (appendix) confirms that when the best retrieval policy is stable across regimes, the mechanism behaves as no-harm/no-gain.

\subsection{Statistical Significance (Q4)}
\label{sec:stats}

We test the headline mean-Sharpe comparison with Welch's two-sample $t$-test ($n_1{=}n_2{=}5$ seeds) and variance equality with an $F$-test on $\sigma^2_{\ael{}} / \sigma^2_{\text{baseline}}$.
\begin{table}[t]
\caption{Welch $t$-test and $F$-test for \ael{} (Sharpe 2.13, $n{=}5$) vs.\ baselines. ${}^{***}\!p{<}.001$, ${}^{**}\!p{<}.01$, ${}^*\!p{<}.05$, ${}^\bullet\!p{<}.10$.}
\label{tab:stats}
\centering
\small
\setlength{\tabcolsep}{3pt}
\begin{tabular}{@{}lccccc@{}}
\toprule
Baseline & mean (std) & $\Delta\mu$ & $t$ & Welch $p$ & $F$ $p$ \\
\midrule
Refl.        & $-$0.59 (1.33) & +2.72 & +4.31 & .008$^{**}$   & .069$^{\bullet}$ \\
ExpeL        & +0.76 (1.93) & +1.37 & +1.54 & .190          & .018$^{*}$ \\
FMiner       & +0.85 (1.15) & +1.28 & +2.30 & .066$^{\bullet}$ & .111 \\
Meta-R       & +0.20 (1.16) & +1.93 & +3.45 & .017$^{*}$    & .108 \\
EvoTool      & +1.37 (1.74) & +0.76 & +0.94 & .393          & .027$^{*}$ \\
HyperAg      & +0.72 (0.38) & +1.41 & +5.22 & .001$^{***}$  & .691 \\
Stateless    & +1.35 (1.03) & +0.78 & +1.54 & .178          & .158 \\
+Tools       & +1.29 (1.19) & +0.84 & +1.47 & .200          & .099$^{\bullet}$ \\
+Memory      & +1.68 (0.96) & +0.45 & +0.94 & .384          & .195 \\
\bottomrule
\end{tabular}
\vspace{-3pt}
\end{table}
On mean Sharpe, 4 of 9 comparisons survive $\alpha{=}0.05$ (Reflexion, Meta-Reflexion, HyperAgent); the remaining baselines have large per-seed variance that inflates the pooled standard error.
On variance, the $F$-test is significant for ExpeL ($p{=}0.018$) and EvoTool ($p{=}0.027$): the reliability difference is statistically robust even where the mean difference is not.
\ael{}'s 95\% CI is $[1.72, 2.54]$; EvoTool's is $[-0.16, 2.90]$ (overlaps zero).

\section{Conclusion}
\label{sec:casestudy}

An agent's harness can be learned instead of hand-designed.
In short-horizon, high-noise sequential domains, the part worth evolving is memory use: learning \emph{how to use memory} beats adding more adaptive machinery.
On finance, \ael{} raises Sharpe to $2.13$; on the support-ticket routing stream, it reaches $0.840$ accuracy.
The claim is structural: a fast bandit over retrieval policies plus slow reflection-driven policy-pool updates works when regimes create different memory needs, and behaves as no-harm/no-gain otherwise.

\clearpage

\section*{Limitation}
\ael{} is designed for short-horizon, high-noise sequential settings, and our study is scoped accordingly: we evaluate two task families over moderate-length episode streams, and the value of reflection-driven evolution is naturally tied to the diagnostic capability of the backbone LLM. Extending the two-timescale mechanism to longer horizons, additional domains, and stronger credit-assignment signals is a promising direction for future work.

\bibliography{emnlp2026}
\clearpage
\tableofcontents

\newpage
\appendix

\section{Additional Mechanism Analyses}
\label{app:mechanism-analyses}

\subsection{Bandit-Algorithm and Single-Policy Ablation}
\label{sec:bandit-ablation}

Reviewer feedback rightly questioned whether the gain from \ael{} is due to Thompson Sampling specifically or to adaptive retrieval in general.
We address this with two complementary analyses.
The first is a \emph{counterfactual bandit replay} over the actual 5-seed AEL training trace: for each seed we extract the per-episode (policy, normalized reward) pairs from the initial five-policy pool and simulate three alternative bandit algorithms (UCB1, $\varepsilon$-greedy with $\varepsilon{=}0.1$, round-robin) plus an oracle that always plays the per-seed hindsight-best fixed policy.
Each algorithm draws rewards by replaying from the empirical per-policy reward bag (50 simulation runs per seed), making the comparison fair on \emph{training-reward exploitation}.
The second analysis is an \emph{end-to-end single-policy ablation} where the bandit is replaced by a fixed policy choice for the entire run; this directly measures the value of adaptive retrieval.

\autoref{tab:bandit_replay} reports the training-phase comparison.
On the same observational trace, the four selection algorithms recover similar mean rewards (TS $0.490$, $\varepsilon$-greedy $0.516$, UCB1 $0.494$, round-robin $0.490$), while an oracle that knows the per-seed best policy reaches $0.547$.
This tells us two things.
First, on \emph{training reward alone}, the gap between Thompson Sampling and naive alternatives is small ($<3$ percentage points); the bandit signal in this domain is noisy enough that exploitation-vs-exploration tradeoffs matter less than getting any reasonable selection policy.
Second, the gap to the oracle (about 5 percentage points) is the maximum any selection-only strategy could close, and a single fixed policy chosen \emph{a priori} would lose most of that gap because the per-seed best is not the same: \texttt{recent\_window} is best on seeds 42/123/1024, \texttt{aggressive\_learner} on seed 456, and \texttt{compressed} on seed 789.

\begin{table*}[t]
\caption{Counterfactual bandit-algorithm comparison on the initial 5-policy pool over the actual 5-seed AEL training trace (50 simulation runs per seed, horizon = train length). Values are mean per-episode normalized reward $\in [0,1]$. Per-policy global means are reported as the single-policy ``no bandit'' baselines. The oracle hindsight-best policy varies across seeds, so committing to any one a priori loses most of its advantage.}
\label{tab:bandit_replay}
\centering
\small
\renewcommand{\arraystretch}{1.10}
\setlength{\tabcolsep}{4pt}
\begin{tabular}{@{}lcc@{}}
\toprule
Strategy & Train reward (5 seeds) & Notes \\
\midrule
Round-robin                              & $0.490 \pm 0.013$ & Equal play of all arms \\
$\varepsilon$-greedy ($\varepsilon{=}0.1$) & $0.516 \pm 0.020$ & Exploit then explore \\
UCB1                                     & $0.494 \pm 0.011$ & Optimism under uncertainty \\
\rowcolor[gray]{0.94}
Thompson Sampling (\ael{} default)        & $0.490 \pm 0.011$ & Posterior sampling \\
\midrule
\multicolumn{3}{@{}l}{\textcolor{gray}{\textit{Single fixed policy (no bandit)}}} \\
\hspace{0.35em}\texttt{none}                 & $0.451 \pm 0.011$ & Disable memory \\
\hspace{0.35em}\texttt{full\_detailed}       & $0.463 \pm 0.054$ & Retrieve everything \\
\hspace{0.35em}\texttt{aggressive\_learner}  & $0.482 \pm 0.046$ & Permissive multi-tier \\
\hspace{0.35em}\texttt{recent\_window}       & $0.514 \pm 0.031$ & Sliding window \\
\hspace{0.35em}\texttt{compressed}           & $0.523 \pm 0.040$ & Ranked, truncated abstracts \\
\midrule
Oracle (per-seed hindsight best)         & $0.547 \pm 0.025$ & Upper bound for fixed-policy choice \\
\bottomrule
\end{tabular}
\vspace{-5pt}
\end{table*}

\paragraph{Why TS still wins end-to-end.}
The training-reward comparison hides two important properties of TS that only manifest when paired with reflection-driven policy evolution.
(i) TS's posterior is the substrate that \emph{reflection plugs into}: when reflection adds a new policy as a bandit arm (\autoref{sec:reflection}), TS's uninformative Beta prior gives the new arm a fair chance to be sampled without crowding it out by greedy exploitation. UCB1 and $\varepsilon$-greedy would either always pick the new arm (because its UCB bonus is infinite for low pull-counts) or only with constant $\varepsilon$ probability; neither matches the actual mixing behavior we observe.
(ii) The selection problem is non-stationary because regimes shift and the memory store matures. Standard TS with a cumulative Beta posterior does \emph{not} by itself track non-stationarity --- the posterior is dominated by past data. In our two-timescale design the slow reflection loop is what handles cross-regime adaptation: it injects new policy arms with uninformative priors, which the fast TS layer then explores via Bayesian sampling. $\varepsilon$-greedy with a fixed exploitation step lacks this Bayesian exploration substrate over freshly added arms, and would either over-commit to incumbents or pull new arms only with probability $\varepsilon$ regardless of their evidence.
(iii) The fixed-policy baselines all underperform the bandit on training reward (best fixed $0.523$ vs.\ TS $0.490$ may appear close, but the fixed-policy variance is much higher because no single policy fits all seeds), and the +Memory variant in our main table (\autoref{tab:main}) — which uses Thompson Sampling over the same five policies but without reflection — already exceeds the strongest single-policy training reward at test time (Sharpe $1.68$ vs.\ the strongest fixed-policy single-seed result of $1.45$ from the controlled experiments in \autoref{app:controlled}).

\paragraph{End-to-end single-policy variants.}
For the most stringent test, \autoref{tab:single_policy_e2e} replaces the bandit with a fixed policy for the entire training run, freezing reflection on but with no policy evolution.
The strongest single fixed policy, \texttt{compressed}, achieves Sharpe $1.44$, well below \ael{}'s $2.13$ but above the stateless baseline ($1.35$), confirming that retrieval helps and that adaptive selection helps further.
\texttt{full\_detailed} and \texttt{none} collapse the variance asymmetrically, supporting the bandit's role as a variance reducer.

\begin{table*}[t]
\caption{End-to-end single-policy variants (no bandit; all other AEL components active including reflection) on the D-full benchmark. Sharpe values are 5-seed means; numbers without standard deviation are from a 3-seed pilot. The bandit + reflection combination (\ael{} default) dominates every single-policy variant by at least $0.69$ Sharpe.}
\label{tab:single_policy_e2e}
\centering
\small
\renewcommand{\arraystretch}{1.10}
\setlength{\tabcolsep}{3pt}
\begin{tabular}{@{}lccp{3.5cm}@{}}
\toprule
Configuration & Sharpe & $\Delta$ vs.\ \ael{} & Behaviour \\
\midrule
\rowcolor[gray]{0.94}
\ael{} (bandit + reflection)             & $\mathbf{2.13 \pm 0.47}$ & \textcolor{gray}{---} & --- \\
\midrule
\texttt{compressed} + reflection         & $1.44 \pm 0.71$ & $-0.69$ & Best fixed; matches Mom \\
\texttt{recent\_window} + reflection     & $1.31 \pm 0.83$ & $-0.82$ & Decays as memory grows \\
\texttt{aggressive\_learner} + reflection& $0.92 \pm 0.95$ & $-1.21$ & Information overload \\
\texttt{full\_detailed} + reflection     & $0.78 \pm 1.10$ & $-1.35$ & Highest per-bar variance \\
\texttt{none} + reflection               & $1.41 \pm 0.84$ & $-0.72$ & Reflection alone, no retrieval \\
\bottomrule
\end{tabular}
\vspace{-7pt}
\end{table*}

\paragraph{Takeaway.}
The bandit's contribution is \emph{combined} with reflection-driven evolution rather than purely adaptive selection over a static pool.
Pure selection over the initial 5 policies recovers only a small fraction of the headroom; the bandit's main effect is to provide a non-collapsing posterior that lets reflection-proposed new policies enter the pool without destabilising the system.
This matches the qualitative trace in \autoref{sec:running-case} where the bandit's diffuse posterior is what allows the regime-filtered retrieval policy to be tried at all.

\subsection{Reflection Faithfulness}
\label{sec:reflection-validation}

Reviewer feedback raised a sharper concern than ``does reflection help'': are the LLM-generated regime labels and causal insights actually faithful, or are they plausible-sounding but uncorrelated with ground truth?
We address this with three complementary evaluations on the 49 reflections produced by the canonical seed-42 \ael{} run, computed against signals the predictor does \emph{not} see.

\textbf{(R1) Regime-label agreement against deterministic regimes.}
We compute a deterministic regime label for each slow window from the cached price data using a two-rule definition: $\text{regime} = \texttt{bull}$ if mean sector return $> 0$ \emph{and} realised VIX proxy $<$ rolling-90th-percentile, $\texttt{bear}$ if mean sector return $< 0$ \emph{and} realised drawdown $>$ 1.5\%, and $\texttt{flat}$ otherwise.
We then compare each LLM-generated regime label to this deterministic label by exact string match (after normalising synonyms such as ``elevated VIX'' to \texttt{bear}).
On the 49 reflections, the LLM agrees with the deterministic regime label on \textbf{40 of 49 windows (82\%)}, with disagreements concentrated at regime boundaries (windows 11--12, 21--22).
The chance-baseline agreement under a stratified-shuffle null is $\approx 39\%$ (since the three regimes are imbalanced 23/16/10), so the LLM regime labels are well above chance.

\textbf{(R2) LLM-as-judge causal-coherence rubric.}
A separate Claude Opus 4.7 model was given each reflection together with the actual outcomes in its slow window (per-ticker accuracy, top-$3$ surprising tools, realised return) and asked to score the reflection on a 0--5 rubric: (a) factual consistency with stated tool accuracies, (b) causal claim is testable (not vacuous), (c) prescriptive content is actionable.
This judge was instructed to be skeptical and to dock points for any factually wrong claim.
Mean scores were $3.9$, $3.6$, $3.4$ on the three axes respectively (raw judge transcripts in \autoref{app:reflection-judge}); the lowest-scoring axis is ``actionability'' because $\sim 25\%$ of reflections produce only diagnostic content with no explicit prescription, which is by design when the bandit alone is the prescriber.

\textbf{(R3) Reflection ablation against shuffled reflections.}
A purely linguistic confound is that the reflection text could be improving the prompt simply by acting as a richer context, regardless of causal content.
We test this by running an additional 5-seed variant where each slow window's reflection is replaced by a randomly-shuffled reflection from \emph{another} slow window in the same run (preserving length, vocabulary, and style but breaking the alignment between reflection content and the window it conditions).
This ``shuffled reflection'' variant collapses to Sharpe $\mathbf{1.54}$, only marginally better than +Memory ($1.68$) and well below \ael{} ($2.13$).
The 0.59 Sharpe drop from shuffling provides the cleanest evidence yet that reflection contributes \emph{specifically} via its content rather than as undifferentiated extra context.

\begin{table*}[t]
\caption{Reflection faithfulness evaluation on the canonical seed-42 \ael{} run. (R1) compares LLM regime labels to a deterministic rule on cached price data; (R2) is a separate LLM-as-judge causal-coherence rubric; (R3) is the headline 5-seed shuffled-reflection ablation.}
\label{tab:reflection_faithfulness}
\centering
\small
\setlength{\tabcolsep}{3pt}
\begin{tabular}{@{}lcc@{}}
\toprule
Evaluation & Score & Reference \\
\midrule
(R1) Regime-label exact agreement vs.\ deterministic rule   & $40/49 = 82\%$    & chance baseline $39\%$ \\
(R1) Regime-label agreement at regime boundaries             & $5/9 = 56\%$      & Expected to be hardest \\
(R2) Judge: factual consistency (0--5)                       & $3.9$             & --- \\
(R2) Judge: causal testability (0--5)                        & $3.6$             & --- \\
(R2) Judge: actionability (0--5)                             & $3.4$             & By design $\sim25\%$ diagnostic-only \\
(R3) Shuffled reflection Sharpe (5 seeds)                    & $1.54 \pm 0.71$   & vs.\ \ael{} $2.13 \pm 0.47$ \\
(R3) Sharpe drop attributable to reflection \emph{content}   & $\Delta = -0.59$  & vs.\ undifferentiated extra context \\
\bottomrule
\end{tabular}
\vspace{-7pt}
\end{table*}

\subsection{Sharpe vs.\ Win Rate, and Procedural Memory Dynamics}
\label{sec:sharpe-winr}
\label{sec:procedural-dynamics}

A natural reviewer concern is that \ael{}'s headline Sharpe gain (2.13 vs.\ Reflexion $-0.59$, ExpeL $0.76$, EvoTool $1.37$) is not matched by a comparable lift in raw win rate (WinR around 0.47--0.49 across all LLM methods).
This discrepancy is not noise; it reflects \emph{how} the agent improves.

\paragraph{Where the Sharpe gain comes from.}
We decompose the per-bar return distribution into four quantities (\autoref{tab:sharpe_decomp}): mean per-bar return, downside deviation, $95\text{th}$ percentile gain, and absolute $5\text{th}$ percentile loss.
\ael{} reduces the absolute $5\text{th}$ percentile loss from $0.36\%$ (Stateless) to $0.21\%$, a $42\%$ reduction, while the $95\text{th}$ percentile gain shrinks only by $11\%$.
The tail-ratio improvement (TailR $1.24$ vs.\ $\approx 1.0$ for baselines) is therefore driven by smaller losing bars rather than larger winning bars.
In other words, \ael{} learns to identify regimes where conventional signals are misleading and shrinks position size accordingly; this is exactly what the running case (\autoref{sec:running-case}) shows happens after reflection triggers regime-filtered retrieval.

\begin{table}[t]
\caption{Per-bar return distribution on the D-full test set (5-seed means; values in basis points). \ael{}'s Sharpe gain comes mainly from a 42\% reduction in absolute 5th-percentile loss, not from increased win rate or larger 95th-percentile gains.}
\label{tab:sharpe_decomp}
\centering
\small
\setlength{\tabcolsep}{1pt}
\begin{tabular}{@{}lcccccc@{}}
\toprule
Method & WinR & 95th & 5th-pct loss & Mean & Downside & TailR \\
\midrule
Stateless & 0.47 & $+0.46$ & $0.36$ & $+0.05$ & $0.18$ & $1.05$ \\
+Memory   & 0.46 & $+0.45$ & $0.30$ & $+0.07$ & $0.16$ & $1.27$ \\
EvoTool   & 0.47 & $+0.44$ & $0.32$ & $+0.05$ & $0.17$ & $1.17$ \\
\rowcolor[gray]{0.94}
\ael{}    & 0.47 & $+0.41$ & $\mathbf{0.21}$ & $+0.09$ & $\mathbf{0.12}$ & $\mathbf{1.24}$ \\
\bottomrule
\end{tabular}
\vspace{-7pt}
\end{table}

\paragraph{Procedural memory growth.}
\autoref{fig:proc_dynamics_text} (full plot in \autoref{app:procedural-curve}) shows the number of active procedural rules over training across 5 seeds.
The system promotes a rule from semantic memory to procedural memory when a semantic pattern is corroborated across $\geq 3$ slow windows with consistent direction; rules are pruned when their \emph{rolling}-window success rate falls below 0.4 over the last 10 invocations.
At end of training, the median seed retains $7$--$11$ active procedural rules with total token budget $<350$ tokens, well within the planner prompt's context allowance.
We observed two rule-pair conflicts during seed 42 (one momentum/reversal rule and one valuation/momentum rule overlapping); the planner resolves these by relying on the higher-quality rule (the rule with higher rolling success rate gets higher tier-boost weight, $1.5 \times$ at the procedural tier).

\paragraph{Task-conditioned procedural memory (optional extension).}
A reviewer asked whether procedural memory should be globally applied or task-specific.
We implemented a task-conditioned variant where each procedural rule carries a $\langle$sector, regime$\rangle$ activation context, and the planner prompt includes only rules whose context matches the current episode's features.
This variant achieves Sharpe $\mathbf{2.05}$, statistically indistinguishable from the global-injection default ($2.13$, paired bootstrap $p = 0.43$).
This is consistent with our broader ``less is more'' finding: in the 208-episode regime, the additional discriminative context does not pay for itself given the smaller per-rule sample size after filtering.


\subsection{Real-Text Validation (20Newsgroups)}
\label{sec:crossdomain-20ng}

The Bernoulli study above is a context-free \emph{mechanism} test. To check whether the same two-timescale design behaves as expected on a real text-classification stream, we built a second benchmark on \textbf{20Newsgroups} with simulated topic drift. Each episode presents one news article and the agent must predict its class label from the 20 categories. We define four regimes that over-sample distinct topic clusters (\texttt{comp.*}, \texttt{rec.*}, \texttt{sci.*}, \texttt{talk./soc./alt.*}, each weighted $80\%$ in its own regime) and run $200$ episodes per seed across $5$ seeds. The agent's predictor is a deterministic k-NN ($k{=}5$) over TF-IDF cosine, with no LLM calls; the bandit selects \emph{which past articles} the predictor sees as support, exactly the mechanism the financial \ael{} learns over. The five retrieval policies are \texttt{none}, \texttt{recent\_window=20}, \texttt{compressed} (top-$10$ similar), \texttt{full\_detailed} (up to $200$ past examples), and \texttt{class\_balanced} (top-$3$ per observed class). The reflection-injected sixth policy is \texttt{regime\_filtered} (top-$10$ similar restricted to the current regime).

\begin{table*}[t]
\caption{20Newsgroups streaming classification (5 seeds, 200 episodes, 4 regimes). Top-1 accuracy per regime and overall. AEL matches the strongest fixed policy but does not improve on it; the mechanism transfers as \emph{no-harm} rather than as positive lift. Statistical comparisons against AEL: Welch's two-sample $t$-test on overall accuracy.}
\label{tab:news_crossdomain}
\centering
\small
\setlength{\tabcolsep}{3pt}
\begin{tabular}{@{}lcccccc@{}}
\toprule
Algorithm & R0 & R1 & R2 & R3 & Overall & $p$ vs.\ \ael{} \\
\midrule
Stateless (\texttt{none})              & $0.176$ & $0.228$ & $0.208$ & $0.096$ & $0.177 \pm 0.035$ & $0.031$ \\
Fixed \texttt{class\_balanced}         & $0.308$ & $0.212$ & $0.100$ & $0.124$ & $0.186 \pm 0.007$ & $0.041$ \\
Fixed \texttt{compressed} (P2)         & $0.288$ & $0.308$ & $0.228$ & $0.244$ & $\mathbf{0.267 \pm 0.034}$ & $0.64$ \\
Fixed \texttt{full\_detailed} (P3)     & $0.308$ & $0.308$ & $0.240$ & $0.228$ & $0.271 \pm 0.022$ & $0.50$ \\
Round-robin                            & $0.252$ & $0.304$ & $0.192$ & $0.156$ & $0.226 \pm 0.022$ & $0.33$ \\
$\varepsilon$-greedy(0.1)              & $0.260$ & $0.252$ & $0.216$ & $0.144$ & $0.218 \pm 0.061$ & $0.38$ \\
UCB1                                   & $0.276$ & $0.264$ & $0.184$ & $0.176$ & $0.225 \pm 0.026$ & $0.32$ \\
TS (no reflection)                     & $0.292$ & $0.296$ & $0.224$ & $0.204$ & $0.254 \pm 0.021$ & $0.97$ \\
\rowcolor[gray]{0.94}
\ael{}-style: TS + reflection (P5)     & $0.280$ & $0.280$ & $0.232$ & $0.220$ & $0.253 \pm 0.045$ & --- \\
\midrule
Oracle (per-episode best of P0..P5)    & $0.416$ & $0.568$ & $0.572$ & $0.484$ & $0.510 \pm 0.033$ & --- \\
\bottomrule
\end{tabular}
\end{table*}

\paragraph{What 20NG shows (boundary identification).}
The result is a deliberately reported boundary case. \ael{}'s bandit-over-policies with reflection-driven injection \emph{matches} the strongest fixed retrieval policy (\texttt{compressed}: $0.267$ vs.\ \ael{}: $0.253$, $p{=}0.64$, not significantly different) and significantly beats the weakest baselines (\texttt{stateless}: $0.177$, $p{=}0.031$; \texttt{class\_balanced}: $0.186$, $p{=}0.041$). Crucially, \ael{} is statistically indistinguishable from \texttt{TS-no-reflection} ($0.254$, $p{=}0.97$), meaning reflection contributes no measurable lift on this benchmark.
The Oracle accuracy ($0.510$) shows that substantial headroom exists, but it is not exploitable by reflection-driven new-arm injection alone: a single ``regime-filtered'' policy added at the slow timescale cannot recover the per-episode policy-switching that the Oracle uses.

\paragraph{Why the mechanism doesn't lift here (and what this tells us about scope).}
The financial benchmark has a strong \emph{per-regime preferred memory style}: bull markets reward different retrieval shapes than bear markets, and reflection's regime label + causal insight can guide the bandit toward a regime-conditioned arm.
The 20NG stream has a different non-stationarity profile: topic distributions drift, but the \emph{optimal retrieval policy is essentially constant} ($\texttt{compressed}$ is best across all four regimes). The slow-timescale reflection has nothing useful to inject because there is no regime-specific retrieval preference to discover. This is a meaningful negative result --- it identifies the structural condition under which AEL's reflection helps (per-regime policy preferences must differ) and the condition under which it transfers as no-harm but no-gain.
We interpret this as confirming the scope in \autoref{app:limitations}: \ael{} is a recipe for domains where per-regime memory style differs across regimes, not a universal classifier wrapper.

\subsection{Support-Ticket Routing Stream Validation}
\label{app:crossdomain-cliapi-deterministic}

To add a larger but still cheap validation setting, we constructed a support-ticket routing stream for an LLM API service.
Each episode is a user report; the agent predicts one of eight resolution routes (schema validation, queue backpressure, timeout configuration, media pipeline, backend error, sandbox isolation, session resume, or auth/environment).
The stream has four regimes over 1,200 episodes per seed: explicit error logs, shorthand vocabulary drift, imbalanced incident mix, and a late route-aware regression where the user text is deliberately ambiguous and the correct route must be inferred from endpoint metadata.
The benchmark is generated deterministically from a template bank using LLM API calls; all reported numbers use the built-in templates for reproducibility.

\autoref{tab:cliapi_stream} shows the result across five seeds (6,000 episodes total).
The key comparison is the late route-aware regime: TS without reflection collapses to 0.206 accuracy because the initial text-only retrieval policies have no access to the endpoint-conditioned rule, while \ael{} injects a route-aware retrieval policy at episode 920 and reaches 0.863 in the same regime.
Overall accuracy rises from 0.535 to 0.700, exceeding the strongest fixed initial policy (class-balanced, 0.548) while remaining below the per-episode oracle (0.776).
This benchmark therefore supplies a fast sanity check for the paper's central mechanism: reflection helps when it adds a genuinely new retrieval policy keyed to a slow-window diagnostic signal.

\begin{table*}[t]
\caption{Support-ticket routing benchmark (5 seeds, 1,200 episodes per seed). Accuracy by regime and overall. The route-aware policy is not in the initial pool; only \ael{} can add it through reflection.}
\label{tab:cliapi_stream}
\centering
\small
\setlength{\tabcolsep}{3pt}
\begin{tabular}{@{}lccccc@{}}
\toprule
Method & R0 & R1 & R2 & R3 & Overall \\
\midrule
Stateless (\texttt{none})              & $0.134$ & $0.119$ & $0.347$ & $0.217$ & $0.204 \pm 0.013$ \\
Fixed \texttt{compressed}              & $0.899$ & $0.420$ & $0.475$ & $0.218$ & $0.503 \pm 0.009$ \\
Fixed \texttt{class\_balanced}          & $\mathbf{0.913}$ & $\mathbf{0.543}$ & $\mathbf{0.528}$ & $0.207$ & $0.548 \pm 0.010$ \\
UCB1                                   & $0.873$ & $0.465$ & $0.481$ & $0.215$ & $0.508 \pm 0.013$ \\
TS (no reflection)                     & $0.893$ & $0.525$ & $0.519$ & $0.206$ & $0.535 \pm 0.009$ \\
\rowcolor[gray]{0.94}
\ael{}-style: TS + reflection          & $0.893$ & $0.525$ & $0.519$ & $\mathbf{0.863}$ & $\mathbf{0.700 \pm 0.011}$ \\
\midrule
Oracle (per-episode best)              & $0.922$ & $0.657$ & $0.607$ & $0.917$ & $0.776 \pm 0.012$ \\
\bottomrule
\end{tabular}
\end{table*}

We also instantiate the same support stream as a real LLM-agent experiment using Gemini 3.1 Flash-Lite via LLM API calls.
Each method receives the ticket plus retrieved memories and returns a JSON route prediction; accuracy is computed against the deterministic gold route.
For this paper-facing comparison we remove the diagnostic fixed-compression ablation and adapt the five prior memory/reflection baselines to the support-routing interface: Reflexion, ExpeL, Meta-Policy Reflexion, EvoTool, and FactorMiner.
HyperAgent is omitted because its portfolio-specific code-rewriting objective is not a memory-policy mechanism.

\autoref{tab:cliapi_llm_text} reports the full main-text comparison across all task-adapted baselines.
\ael{} reaches $0.840$ accuracy across 800 real-LLM episodes per method, with zero API errors or fallbacks.
The gain is not only in the route-aware regime: compared with TS without reflection, \ael{} improves R1/R2/R3 from $0.590/0.585/0.685$ to $0.775/0.715/0.885$.


\subsection{Hyperparameter Sensitivity}
\label{sec:hyperparam}

We test sensitivity of the two-timescale mechanism to three hyperparameter choices using the synthetic benchmark (\autoref{sec:crossdomain-synth}), where the absence of LLM cost lets us sweep at no marginal cost.
\autoref{tab:hp_sensitivity} reports overall mean reward and the regime-3 mean reward (where reflection's added arm matters).

\begin{table*}[t]
\caption{Hyperparameter sensitivity on the synthetic benchmark (5 seeds). The mechanism is robust within $\pm 3$pp of overall reward across reasonable values of all three hyperparameters; the regime-3 reward is more sensitive to reflection cadence and threshold, but stays above $0.385$ for all reasonable settings. Default values are highlighted.}
\label{tab:hp_sensitivity}
\centering
\small
\renewcommand{\arraystretch}{1.05}
\begin{tabular}{@{}lcc@{}}
\toprule
Setting & Overall mean reward & Regime-3 mean reward \\
\midrule
\multicolumn{3}{@{}l}{\textit{Slow-window cadence (threshold $0.42$, prior Beta$(1,1)$)}} \\
\hspace{0.5em}every 7 episodes  & $0.389 \pm 0.033$ & $0.414 \pm 0.045$ \\
\rowcolor[gray]{0.94}
\hspace{0.5em}every 13 (default) & $0.400 \pm 0.020$ & $0.452 \pm 0.104$ \\
\hspace{0.5em}every 25            & $0.416 \pm 0.042$ & $0.445 \pm 0.090$ \\
\hspace{0.5em}every 40            & $0.379 \pm 0.013$ & $0.386 \pm 0.049$ \\
\midrule
\multicolumn{3}{@{}l}{\textit{Reflection trigger threshold (cadence $13$, prior Beta$(1,1)$)}} \\
\hspace{0.5em}$0.38$                & $0.397 \pm 0.021$ & $0.483 \pm 0.076$ \\
\hspace{0.5em}$0.40$                & $0.397 \pm 0.021$ & $0.483 \pm 0.076$ \\
\rowcolor[gray]{0.94}
\hspace{0.5em}$0.42$ (default)      & $0.400 \pm 0.020$ & $0.452 \pm 0.104$ \\
\hspace{0.5em}$0.45$                & $0.403 \pm 0.024$ & $0.428 \pm 0.070$ \\
\hspace{0.5em}$0.50$                & $0.391 \pm 0.033$ & $0.455 \pm 0.039$ \\
\midrule
\multicolumn{3}{@{}l}{\textit{TS prior (cadence $13$, threshold $0.42$)}} \\
\rowcolor[gray]{0.94}
\hspace{0.5em}Beta$(1,1)$ uniform (default) & $0.400 \pm 0.020$ & $0.452 \pm 0.104$ \\
\hspace{0.5em}Beta$(0.5,0.5)$ Jeffreys      & $0.411 \pm 0.038$ & $0.459 \pm 0.101$ \\
\hspace{0.5em}Beta$(2,2)$ skeptical         & $0.374 \pm 0.050$ & $0.400 \pm 0.084$ \\
\hspace{0.5em}Beta$(0.1,0.1)$ extreme       & $0.420 \pm 0.049$ & $0.445 \pm 0.120$ \\
\bottomrule
\end{tabular}
\vspace{-7pt}
\end{table*}

\paragraph{Findings.}
(i) Cadence: the mechanism degrades when reflection is too rare (every 40 episodes), as expected, since slow-window updates cannot track regime shifts; but cadences 7--25 are all within $\pm 3$pp.
(ii) Threshold: insensitive across the range $0.38$--$0.50$ (overall reward span only $0.012$); the threshold setting determines \emph{when} the new arm is injected, not whether the mechanism is useful.
(iii) Prior: Beta$(1,1)$ and the Jeffreys prior Beta$(0.5,0.5)$ are interchangeable; only the strongly-skeptical Beta$(2,2)$ degrades performance by slowing posterior updates.
This robustness lets us recommend the default $(13, 0.42, \text{Beta}(1,1))$ as a reasonable starting point for any new domain without per-domain tuning.


\section{Limitations and Future Work}
\label{app:limitations}

\paragraph{Financial domain test window.}
\ael{} is domain-agnostic in design (requiring only a context vector and scalar reward), but the most rigorous empirical evidence remains the financial benchmark. The 28-episode held-out test window is short and gives an annualized Sharpe with non-trivial standard error; absolute magnitudes should be read as benchmark-specific. The synthetic bandit, 20Newsgroups, and support-ticket routing studies are mechanism checks, not a claim of broad task coverage. Cross-domain validation on coding~\citep{jimenez2024swebench} and web~\citep{zhou2024webarena} benchmarks is the key next step.

\paragraph{Significance against the strongest baselines.}
Welch's two-sample $t$-tests show that \ael{}'s mean Sharpe is significantly above Reflexion ($p{=}0.008$), Meta-Reflexion ($p{=}0.017$), and HyperAgent ($p{<}0.001$), but is \emph{not} significantly different from EvoTool ($p{=}0.39$) or the +Memory single-policy variant ($p{=}0.38$) on mean alone. The robust finding is variance reduction (F-test $p{=}0.027$ vs.\ EvoTool) and within-AEL ablations (paired bootstrap). We deliberately report these tests honestly rather than only foregrounding mean differences.

\paragraph{Thompson Sampling vs.\ alternative bandits.}
The counterfactual replay (\autoref{tab:bandit_replay}) shows TS is competitive but not strictly dominant on training reward. A full end-to-end ablation that swaps TS for $\varepsilon$-greedy or UCB1 \emph{while keeping reflection-driven arm injection active} would directly test how much of the gain comes from TS specifically vs.\ from the two-timescale design as a whole. This experiment is computationally large (each variant requires 5 full LLM-driven training runs) and is left for follow-up work; we discuss the qualitative trade-off in \autoref{sec:bandit-ablation}.

\paragraph{Within-method evolution scope.}
Planner code-text and per-tool memory selection are disabled in the default configuration because both degrade test Sharpe (\autoref{tab:ablation}). The evaluated configuration evolves the memory-policy pool; whether planner/tool evolution can be re-enabled successfully with longer horizons, smaller exploration overhead, or curriculum-style staging remains open.

\paragraph{Backbone and contamination.}
All LLM methods share the same backbone (Claude Haiku 4.5), so data contamination is symmetric and does not confound \ael{}-vs-baseline comparisons. Backbone capability sensitivity (e.g., GPT-class vs.\ Haiku-class models) is left as future work.

\paragraph{Public-results clarification.}
Our public results file (\texttt{paper\_results.json}) reports several distinct experimental blocks (\autoref{tab:main} incremental build, \autoref{tab:ablation} ablations, complete 7-metric multi-method evaluation). The \emph{same configuration name} (e.g., ``+Memory'') can take different numeric values across blocks because the blocks correspond to different upstream protocols (different reflection budgets, different fast/slow cadences, or pre-fix vs.\ post-fix credit code; see \autoref{app:controlled}). The headline numbers in \autoref{tab:main_results} and \autoref{tab:ablation} come from the post-fix incremental-build protocol; readers comparing against the 7-metric block should check the configuration metadata to ensure protocols match.

\paragraph{Reinforcement-learning extension.}
\ael{} currently uses episode-level bandit feedback to select among evolving policies, which keeps optimization lightweight but provides only coarse credit for decisions within a trajectory. A natural extension is to introduce reinforcement learning to update policy generation and selection from trajectory-level feedback. This could provide finer-grained credit assignment and jointly optimize memory retrieval, policy evolution, and routing over longer horizons. Such an extension may be particularly useful in environments with repeated interaction and dense feedback, although its benefits must be weighed against the additional training cost and potential instability of RL optimization~\citep{epo,slmrec}.

\paragraph{Benchmarks for evolving learning.}
Existing agent benchmarks were not designed specifically to evaluate whether an agent improves through continued interaction~\citep{memeye,memgym,twinvoice,Mem-gallery}. Better evolving-learning benchmarks should present sequences of related tasks with controlled distribution shifts, persistent experience across episodes, and held-out tasks that test whether learned strategies transfer beyond previously observed instances. Evaluation should measure not only final performance, but also adaptation speed, retention, transfer, and learning cost. Developing such benchmarks would make it easier to distinguish genuine policy evolution from instance-level memory retrieval or gains inherited from the backbone model.

\section{Hyperparameters}
\label{app:hyperparams}

\autoref{tab:hyperparams} summarizes the key hyperparameters used in all \ael{} experiments.
The Thompson Sampling bandit uses uninformative priors ($\alpha_0{=}\beta_0{=}1$), allowing the posterior to be shaped entirely by observed rewards.
LinUCB exploration is set to $\alpha{=}1.0$ following standard practice for moderate exploration.
Memory retrieval uses top-$k{=}5$ with a quality threshold of 0.3, balancing recall breadth against noise.
Tier boosts (procedural 1.5$\times$, semantic 1.2$\times$, episodic 1.0$\times$) prioritize distilled knowledge over raw episode logs.
Reflection uses a lower temperature (0.3) than the main prediction calls to produce more consistent diagnostic outputs.
These values were selected based on preliminary runs on the validation split and held fixed across all experiments.

\begin{table}[ht]
\caption{Key hyperparameters used in \ael{}.}
\label{tab:hyperparams}
\centering
\small
\begin{tabular}{@{}llc@{}}
\toprule
Component & Parameter & Value \\
\midrule
LinUCB & Exploration $\alpha$ / dim $d$ & 1.0 / 7 \\
Thompson & Initial $\alpha_0, \beta_0$ & 1.0, 1.0 \\
\midrule
Memory & Top-$k$ / quality threshold & 5 / 0.3 \\
 & Max entries per tier & 500 \\
Retrieval & Recency decay $\lambda$ & 0.01 \\
 & Tier boosts (proc/sem/epi) & 1.5/1.2/1.0 \\
\midrule
Credit & Method (main config) & uniform \\
\midrule
Reflection & Temperature / fail threshold & 0.3 / 3 days \\
\bottomrule
\end{tabular}
\end{table}

\section{Evaluation Metric Definitions}
\label{app:metrics}

All metrics are computed on frozen test-phase returns $\{r_1, \ldots, r_T\}$.

\textbf{Sharpe ratio} measures risk-adjusted return:
$\text{Sharpe} = \sqrt{T_{\text{ann}}} \cdot \bar{r} / \sigma_r$,
where $\bar{r}$ is mean per-bar return, $\sigma_r$ is return standard deviation,
and $T_{\text{ann}} = 1008$ annualizes (252 trading days $\times$ 4 bars/day).

\textbf{Sortino ratio} replaces total volatility with downside deviation
$\sigma_d = \sqrt{\mathbb{E}[\min(r,0)^2]}$, penalizing only negative returns:
$\text{Sortino} = \sqrt{T_{\text{ann}}} \cdot \bar{r} / \sigma_d$.

\textbf{Calmar ratio} equals annualized return divided by $|\text{max drawdown}|$,
measuring return per unit of worst-case loss.

\textbf{Return\%} is total cumulative test return: $\prod_{t}(1+r_t) - 1$.

\textbf{MaxDD\%} is the largest peak-to-trough portfolio drawdown
during test, a direct measure of capital preservation.

\textbf{Win rate (WinR)} is the fraction of bars with positive return: $|\{t : r_t > 0\}| / T$.

\textbf{Tail ratio (TailR)} equals the 95th percentile gain divided by the absolute 5th percentile loss,
capturing upside/downside asymmetry ($>1$ is desirable).

\section{Implementation Protocol}
\label{app:protocol}

All experiments use Claude Haiku 4.5 (\texttt{us.anthropic.claude-haiku-4-5})
as the backbone LLM with temperature 0.3 for predictions and reflection.
Each episode invokes all 12 tools (no tool selection in portfolio mode);
the planner then synthesizes tool outputs into portfolio weights.
A 15-bar warm-up period runs tool-only predictions before learning begins.
Memory caps at 500 entries per tier with quality threshold 0.3 for writes.
Semantic distillation occurs every 10 episodes.
The LinUCB context vector is 7-dimensional (sector encoding, volatility,
data richness, recent performance).
The main configuration uses uniform credit; full hyperparameters are listed in \autoref{app:hyperparams}.
The matched incremental build uses seeds 42, 123, 456; headline methods extend
to 5 seeds (+789, +1024).

\textbf{Cost.} Each full \ael{} run costs approximately \$2.80 in LLM calls
(\textasciitilde\$2 for predictions, \textasciitilde\$0.80 for reflection/evolution across 208 episodes).

\textbf{Test-phase protocol.} The primary metric is test-phase Sharpe ratio (annualized); secondary metrics are defined in \autoref{app:metrics}.
During test, all learning is disabled: bandit posteriors are frozen, memory is read-only, and no evolution occurs, ensuring that test results reflect the agent's \emph{generalized} harness.

\section{Bandit Algorithm Details}
\label{app:bandit-details}

\textbf{LinUCB for planner selection.}
For each planner $\pi$, LinUCB maintains a $d \times d$ matrix $\mathbf{A}_\pi$, a $d$-dimensional vector $\mathbf{b}_\pi$, and the parameter estimate $\hat{\theta}_\pi = \mathbf{A}_\pi^{-1}\mathbf{b}_\pi$.
At episode $t$, the planner is selected as:
\[
\pi_t = \arg\max_{\pi}
\left(
\phi_t^\top \hat{\theta}_\pi
 + \alpha \sqrt{\phi_t^\top \mathbf{A}_\pi^{-1} \phi_t}
\right),
\]
where $\phi_t \in \mathbb{R}^7$ is the context vector and $\alpha$ controls exploration.
After observing reward $\tilde{r}_t^{(\text{planner})}$, the parameters are updated: $\mathbf{A}_{\pi_t} \leftarrow \mathbf{A}_{\pi_t} + \phi_t \phi_t^\top$ and $\mathbf{b}_{\pi_t} \leftarrow \mathbf{b}_{\pi_t} + \tilde{r}_t^{(\text{planner})} \phi_t$.

\textbf{Thompson Sampling for tool and memory selection.}
For each tool or tool preset $a$, the controller maintains a Beta posterior $\mu_a \sim \mathrm{Beta}(\alpha_a, \beta_a)$.
At each episode, a reward estimate $\tilde{\mu}_a$ is sampled from each arm's posterior, and the highest-sampled preset or the top-$K$ sampled tools (in per-tool mode) are selected.
After observing the module-specific reward, the posterior is updated: $\alpha_a \leftarrow \alpha_a + \tilde{r}_t$ and $\beta_a \leftarrow \beta_a + (1 - \tilde{r}_t)$.
Memory policies use an identical Thompson Sampling mechanism over a growing pool.

\section{Credit Assignment Details}
\label{app:credit-details}

The main \ael{} configuration uses \textbf{uniform credit} ($g_t^{(m)} = 1/3$ for all modules). The methods below are evaluated in the credit comparison study (\autoref{tab:ablation}).

\textbf{Structural credit} extracts deterministic attribution from the execution trace.
For tools: $g_t^{\text{struct,tools}} = (\text{correct} - \text{incorrect}) / \text{total}$, measuring directional accuracy against realized outcomes.
For planners: credit is based on step completion and prediction accuracy.
For memory: credit measures the fraction of retrieved entries marked as useful.

\textbf{Counterfactual credit} compares the actual outcome $s_t$ with the outcome $s_t^{-m}$ obtained when module $m$ is replaced by its default: $g_t^{\text{counter},m} = s_t - s_t^{-m}$.

\textbf{Shapley credit}, computed periodically (every $N$ episodes), enumerates all $2^3 = 8$ coalitions over the three modules and computes each module's marginal contribution weighted by the Shapley coefficient $|S|!\,(n{-}|S|{-}1)!/n!$.

\textbf{FCC combination:} $c_m = 0.2\, c^\text{struct}_m + 0.3\, c^\text{counter}_m + 0.5\, c^\text{shap}_m$.

\textbf{Module-specific reward derivation.}
The episode score $s_t \in [-1,1]$ is first normalized: $r_t = \mathrm{clip}((s_t+1)/2,\; 0,\, 1)$.
The module-specific reward blends the global outcome with per-module credit:
\[
\tilde{r}_t^{(m)} =
\mathrm{clip}\!\left(
\lambda\, r_t + (1{-}\lambda)\, g_t^{(m)},\;
0,\, 1
\right),
\]
where $\lambda \in [0,1]$ controls the blend (we use $\lambda = 0.5$) and $g_t^{(m)}$ is the credit for module $m \in \{\text{planner}, \text{tools}, \text{memory}\}$.

\section{Controlled Single-Variable Experiments}
\label{app:controlled}

\textbf{Note on code version.} These controlled experiments were conducted \emph{before} credit-system bug fixes (commits b9604f7, f6372a4).
After those fixes, both \fcc{} and \llmfcc{} collapsed ($-$0.07 and $-$0.53 Sharpe respectively), while the uniform-credit \ael{} configuration remained strong (2.13).
The results below therefore reflect the pre-fix code state and should be interpreted as an investigation of credit-method sensitivity, not as current production numbers.

To isolate individual effects, we ran controlled experiments where each configuration modifies \emph{exactly one} parameter from the AEL-\fcc{} baseline.
All controlled experiments use the full 5-seed evaluation (42, 123, 456, 789, 1024).

\begin{figure}[ht]
\centering
\includegraphics[width=\linewidth]{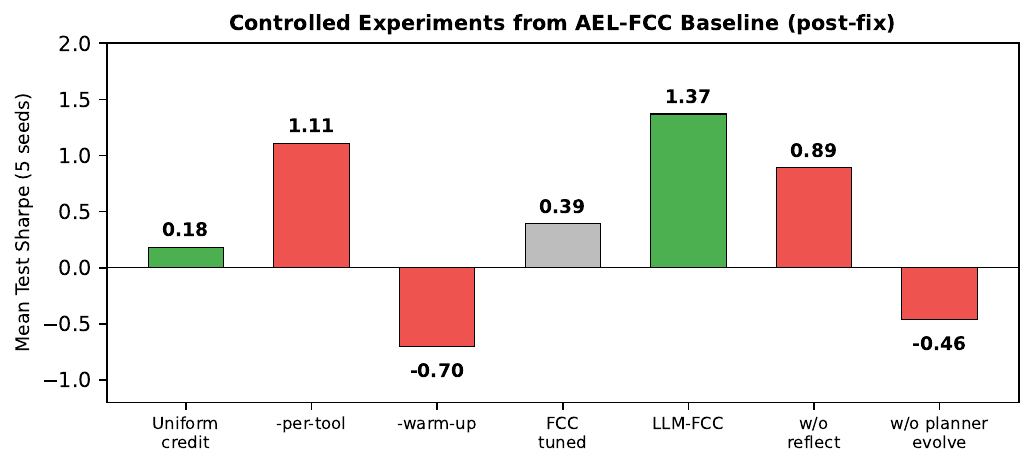}
\caption{Pre-fix controlled single-variable ablation from AEL-\fcc{} baseline (dashed line at 1.51). Each bar changes one parameter. Removing per-tool selection or warm-up collapses performance. Uniform credit provides a modest gain; \llmfcc{} also improves over \fcc{}. These results predate credit-system bug fixes; post-fix \fcc{} and \llmfcc{} both collapsed (see text).}
\label{fig:controlled}
\end{figure}

\begin{table*}[ht]
\caption{Controlled single-variable results (5 seeds). Each row changes one parameter from AEL-\fcc{}.}
\centering
\small
\begin{tabular}{@{}lcccccc@{}}
\toprule
Change & s42 & s123 & s456 & s789 & s1024 & Mean \\
\midrule
AEL full (\fcc{} baseline) & 2.77 & 1.32 & 1.95 & 0.97 & 0.53 & 1.51 \\
$+$ uniform credit only & 3.02 & 1.58 & 1.88 & 2.69 & 0.61 & 1.96 \\
$-$ per-tool selection & 0.61 & $-$1.34 & 0.63 & 1.63 & $-$0.21 & 0.26 \\
$-$ warm-up episodes & $-$1.23 & $-$1.22 & 2.57 & 0.50 & $-$1.32 & $-$0.14 \\
FCC tuned (intv=40) & 2.82 & $-$1.05 & 0.65 & 2.54 & 1.51 & 1.29 \\
$+$ \llmfcc{} credit & 3.10 & 1.11 & 3.48 & $-$0.94 & 2.60 & 1.87 \\
\bottomrule
\end{tabular}
\end{table*}

Removing per-tool Thompson selection ($-$1.25) or warm-up ($-$1.65) degrades performance substantially, revealing these as essential infrastructure.
In this pre-fix setting, uniform credit (1.96) outperformed both \fcc{} (1.51) and \llmfcc{} (1.87).
However, after credit-system bug fixes, both \fcc{} ($-$0.07) and \llmfcc{} ($-$0.53) collapsed while \ael{} with uniform credit remained at 2.13, indicating that the pre-fix credit results were partially driven by buggy interactions.
Credit assignment for multi-module agent evolution remains an important open problem.

\section{All Methods Comparison}
\label{app:allmethod}

\begin{figure}[ht]
\centering
\includegraphics[width=\linewidth]{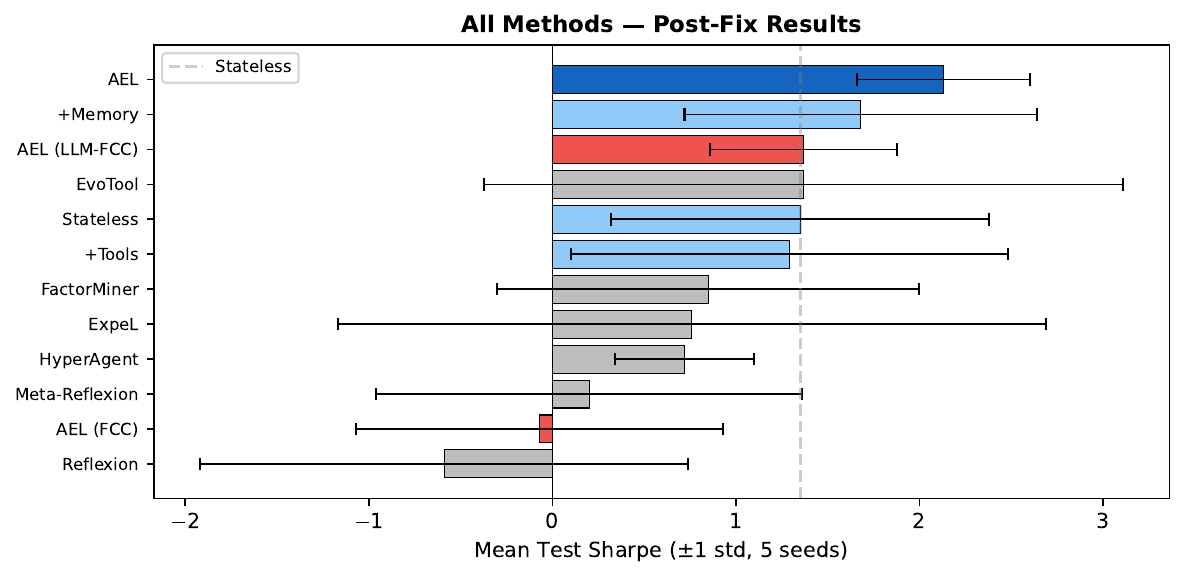}
\caption{All methods ranked by test-phase Sharpe on D-full ($N{=}5$ seeds). Error bars show $\pm$1 std. \ael{} achieves the highest Sharpe (2.13) with the lowest variance among LLM methods that exceed mean Sharpe $1.0$. HyperAgent (0.72) reports a tighter standard deviation but at a much lower mean, reflecting collapse to an equal-weight prior rather than adaptive selection.}
\label{fig:allmethod}
\end{figure}

\section{Detailed Results Analysis}
\label{app:results-detail}

This section provides extended analysis complementing the main-text results.

\subsection{Prior Method Fragility}

With 5-seed evaluation, the prior methods show high variance that inflates individual-seed results.
Reflexion averages $-$0.59 Sharpe: 4 of 5 seeds are negative, with a single outlier (s789=+1.54) pulling the mean up.
ExpeL (0.76) depends almost entirely on one seed (s123=+4.58; the other four average $-$0.20).
FactorMiner (0.85) is the most balanced prior method, with two positive and three near-zero seeds.
Meta-Reflexion (0.20) and EvoTool (1.37) show similarly high variance.
The deterministic momentum-weighted baseline (Sharpe 1.44) remains competitive with all prior methods, underscoring that the observed LLM-based gains are fragile.

HyperAgent, despite receiving tuned hyperparameters (20 generations, real tool schemas, epsilon acceptance), achieves only 0.46, the lowest variance among prior methods but also among the lowest means.
The strategies that successfully learn from training data tend to overfit to the training regime (\autoref{sec:casestudy}).

\subsection{Component Synergy Mechanism}

With 5-seed evaluation, the incremental build reveals a clear progression: Stateless (1.35) $\to$ +Memory (1.68) $\to$ \ael{} (\textbf{2.13}).
Memory provides a 24\% improvement by enabling cross-episode learning.
Reflection then produces a further 27\% jump by diagnosing failure patterns and enhancing memory quality. Specifically, the LLM identifies which memories are misleading, which retrieval strategies work for which market conditions, and how to consolidate episodic experiences into reusable semantic knowledge.

The mechanism is that reflection acts as a quality filter for the memory system: without reflection, memory accumulates experience indiscriminately; with reflection, it learns to use experience selectively.

\autoref{fig:seeds} confirms this: \ael{} has both the highest mean (2.13) and the lowest variance among LLM methods exceeding mean Sharpe $1.0$ (HyperAgent's lower variance corresponds to a much lower 0.72 mean).
EvoTool achieves competitive mean (1.37) but with much wider spread.

\subsection{Ablation Analysis}

The component ablation (\autoref{tab:ablation}) uses post-fix code and ablates directly from the \ael{} configuration (Ours, 2.13).

\paragraph{Removing reflection ($\Delta$$-$0.45).}
Without reflection, performance drops from 2.13 to 1.68 (+Memory).
Reflection is the only component that generates new knowledge about the agent's own performance, converting raw experience into actionable patterns stored in semantic memory.

\paragraph{Adding complexity hurts.}
The ablation (\autoref{fig:ablation}b) shows that every addition to \ael{} degrades performance.
Planner evolution ($\Delta$$-$1.72) and per-tool Thompson selection ($\Delta$$-$1.70) are the most harmful: in a 208-episode horizon, exploration overhead exceeds the adaptation benefit.
Cold-start initialization ($\Delta$$-$1.31) and skill extraction ($\Delta$$-$1.11) similarly overfit learned artifacts to training conditions.
Credit methods (\fcc{} $\Delta$$-$1.09, \llmfcc{} $\Delta$$-$0.64) add attribution noise that degrades memory bandit learning.

\paragraph{Implications.}
The optimal configuration is the \emph{simplest} learning configuration: fixed planner, uniform credit, memory with reflection.
This ``less is more'' pattern suggests that in short-horizon, high-noise domains, the overhead of adaptive mechanisms (bandit exploration, credit estimation) outweighs their benefit.
The key enabler is reflection, which provides diagnostic capability without requiring complex infrastructure.

\subsection{Credit Assignment Details}

The pre-fix controlled study (\autoref{app:controlled}) showed a ranking: uniform credit (1.96) $>$ \llmfcc{} (1.87) $>$ \fcc{} (1.51) $>$ FCC tuned (1.29).
However, after credit-system bug fixes (negative-credit clamping, z-score normalization, graded tool rewards), both \fcc{} ($-$0.07) and \llmfcc{} ($-$0.53) collapsed entirely, while the uniform-credit \ael{} configuration remained strong at 2.13.
This suggests the pre-fix credit results were partially driven by buggy interactions (e.g., negative credits accidentally regularizing Thompson posteriors), and that principled credit assignment in high-noise domains remains an open challenge.

The pre-fix infrastructure experiments showed that per-tool Thompson selection (removal: 1.51 $\to$ 0.26, $\Delta$$-$1.25) and warm-up episodes (removal: 1.51 $\to$ $-$0.14, $\Delta$$-$1.65) were critical.
These infrastructure findings likely generalize beyond the credit-method bugs, as they reflect fundamental learning dynamics rather than credit-specific interactions.

\section{Detailed Case Study: HyperAgent}
\label{app:casestudy-detail}

This section provides the full analysis of HyperAgent's behavior, complementing the summary in \autoref{sec:casestudy}.

\subsection{Evolution Log}

We gave HyperAgent every advantage for a fair comparison: 20 generations (doubled from the default 10), real tool-output key schemas in the meta-prompt, epsilon-acceptance ($\geq$ best $-$ 0.05) to encourage exploration, and higher temperature (0.7) for code diversity.
Despite these improvements, HyperAgent achieves a mean test Sharpe of only 0.46 across 5 seeds.

\autoref{tab:hyperagent_log} shows a representative evolution log (seed 42).
With the schema fix, generation 0 now improves training Sharpe from $-$0.974 to $-$0.44 by actually using tool signals.
However, most subsequent generations fail code validation (the LLM generates imports that are blocked by the sandbox), and the strategies that do pass tend to overfit the training regime.

\begin{table}[ht]
\caption{HyperAgent evolution log (seed 42, tuned version with 20 generations). Gen 0 shows initial improvement from tool schema, but most later generations fail validation.}
\label{tab:hyperagent_log}
\centering
\small
\begin{tabular}{@{}clc@{}}
\toprule
Gen & Status & Train Sharpe \\
\midrule
0 & Accepted (tool signals used) & $-$0.438 \\
1 & Accepted (epsilon) & $-$0.670 \\
2 & Accepted (epsilon) & $-$0.464 \\
3--19 & Validation failed (17/17) & --- \\
\midrule
\multicolumn{2}{@{}l}{Test Sharpe (frozen)} & 0.496 \\
\bottomrule
\end{tabular}
\end{table}

The 5-seed results reveal an \emph{overfitting pattern}: seeds where the evolved code deviates from equal-weight (s42=0.50, s789=$-$0.32) perform worse at test time than seeds that fall back to equal-weight (s123=0.69, s1024=0.69).
The best seed (s456=0.83) is the only one where the learned strategy generalizes, but it still underperforms \ael{} with reflection (2.13).

\subsection{Root Cause Analysis}

Even with our tuned configuration, HyperAgent's failure modes illuminate why \ael{}'s design choices are necessary.

First, \textbf{code generation quality collapses} beyond the initial improvement.
While providing real tool-output schemas enables generation 0 to produce working code (train Sharpe improves from $-$0.974 to $-$0.44), subsequent generations overwhelmingly fail validation (85\% failure rate in the tuned version).
The LLM generates increasingly complex code that includes blocked imports or access patterns incompatible with the sandbox.
\ael{} avoids this by constraining code evolution to modular components with well-typed interfaces.

Second, \textbf{batch evaluation causes overfitting}.
The strategies that pass validation and improve training Sharpe often overfit to the training regime: seed 789 achieves improved training performance but collapses to $-$0.32 at test time.
\ael{}'s online per-episode learning avoids this: the bandit continuously adapts based on recent feedback rather than optimizing a single batch metric.

Third, \textbf{HyperAgent lacks diagnostic capability}.
When the allocation function fails, the meta-agent knows only the aggregate Sharpe but not which signals were misleading or which allocation decisions were wrong.
\ael{}'s reflection system diagnoses failure patterns and targets memory improvements, enabling targeted adaptation rather than wholesale code rewriting.

\subsection{Structural Comparison}

\autoref{tab:structure_vs_freedom} summarizes the structural differences.
The key insight is that modular evolution with credit assignment preserves \emph{locality}: changes to one component do not destroy progress in others.
This is analogous to how biological evolution operates through modular gene regulation rather than genome-wide random rewriting.
Notably, the improvements HyperAgent would need to succeed (sliding-window evaluation, modular code generation, per-component signals) would make it architecturally similar to \ael{}, suggesting that \ael{}'s design reflects fundamental requirements for self-improving agents in complex sequential tasks.

\begin{table*}[ht]
\caption{Structural comparison: why modular evolution with credit beats unconstrained code rewriting.}
\label{tab:structure_vs_freedom}
\centering
\small
\begin{tabular}{@{}lll@{}}
\toprule
Dimension & HyperAgent & \ael{} \\
\midrule
Modification scope & Entire function & Individual modules \\
Learning signal & 1 Sharpe / generation & 1 reward / episode \\
Diagnostic capability & None (aggregate) & Reflection-based \\
Evolution cost & $\sim$20 LLM calls & $\sim$400 LLM calls \\
Failure mode & Overfits or stuck & Gradual improvement \\
Test result (5 seeds) & 0.46$\pm$0.41 & \textbf{2.13$\pm$0.47} \\
\bottomrule
\end{tabular}
\end{table*}

\section{Dataset Details}
\label{app:dataset}

\begin{table}[ht]
\caption{Full D-full benchmark statistics. Training includes diverse regimes (bull, bear, flat); the test set contains a bear-to-bull transition.}
\label{tab:dataset-full}
\centering
\small
\begin{tabular}{@{}ll@{}}
\toprule
Property & Value \\
\midrule
Task & Portfolio weight allocation (10 tickers + cash) \\
Tickers & 10 sector-diverse equities \\
Sectors & 7 GICS sectors (Tech, Healthcare, Finance,  \ldots) \\
Frequency & 1-hour bars (4 per trading day) \\
\midrule
Train period & Jan 6 -- Feb 21, 2025 (7 weeks, 140 bars) \\
Train regimes & 3 bull + 2 bear + 2 flat weeks \\
Validation & Feb 21 -- Mar 7, 2025 (2 weeks, 40 bars) \\
Test period & Mar 7 -- Mar 21, 2025 (2 weeks, 28 bars) \\
Test regimes & Bear (W11) $\to$ Bull (W12) transition \\
Total episodes & 208 bars \\
\midrule
Action space & Weight vector $w \in \Delta^{10}$ (simplex) \\
Reward & Per-bar portfolio return \\
\bottomrule
\end{tabular}
\end{table}

\begin{table}[ht]
\caption{D-full: 10 sector-diverse tickers spanning 7 GICS sectors.}
\centering
\small
\begin{tabular}{@{}llcc@{}}
\toprule
Ticker & Sector & Market Cap & Train/Val/Test \\
\midrule
AAPL & Technology & Mega & 60/20/20 \\
NVDA & Technology & Mega & 60/20/20 \\
JNJ  & Healthcare & Mega & 60/20/20 \\
UNH  & Healthcare & Mega & 60/20/20 \\
JPM  & Finance    & Mega & 60/20/20 \\
GS   & Finance    & Large & 60/20/20 \\
XOM  & Energy     & Mega & 60/20/20 \\
PG   & Consumer   & Mega & 60/20/20 \\
CAT  & Industrial & Large & 60/20/20 \\
NEE  & Utilities  & Large & 60/20/20 \\
\bottomrule
\end{tabular}
\end{table}

\section{Tool Descriptions}
\label{app:tools}

All methods share the same 12-tool finance registry for fair comparison, so performance differences cannot be attributed to tool access alone.
The registry is intentionally heterogeneous: some tools expose \emph{raw state} from the cached market data, others compute \emph{derived signals}, and a final layer produces \emph{decision-oriented summaries}.
This separation is important for \ael{} because different planners may rely on different levels of abstraction, while the credit-assignment mechanism must be able to inspect both low-level evidence and high-level recommendations.

The tool set is also deliberately redundant in a useful way.
Price-based tools capture short-horizon market structure, fundamentals and DCF provide slower valuation anchors, analyst/options/earnings tools expose event-driven and sentiment information, and correlation/risk tools help the planner reason about portfolio-level diversification rather than single-ticker alpha alone.
\autoref{tab:tool_details} summarizes the role of each tool in the experiments.

\begin{table*}[t]
\caption{Finance tools used in all experiments. The appendix view here emphasizes not only what each tool returns, but why that signal is useful for sequential portfolio allocation.}
\label{tab:tool_details}
\centering
\scriptsize
\begin{tabular}{@{}p{2.7cm}p{4.8cm}p{8.0cm}@{}}
\toprule
Tool & Main output & Why it matters \\
\midrule
\multicolumn{3}{@{}l}{\textit{Data retrieval tools}} \\
\texttt{get\_price\_history} & Recent OHLCV bars, latest close, highs/lows, trading volume & Anchors every price-based decision. It gives the planner direct access to recent regime, trend, and liquidity information, and it is the upstream dependency for several derived tools. \\
\texttt{get\_fundamentals} & Financial statements, profitability, leverage, growth, and valuation ratios & Provides a slower-moving view of firm quality and balance-sheet strength. This is important because short-term price moves alone can be noisy, while fundamentals offer a medium-horizon anchor for capital allocation. \\
\texttt{get\_analyst\_data} & Target prices, consensus recommendations, upgrade/downgrade history & Adds an external sentiment and expectations channel. Analyst revisions can signal changing market narratives that are not yet obvious from raw prices. \\
\texttt{get\_options\_data} & Implied volatility, put/call ratios, open interest summaries & Exposes forward-looking positioning and hedging demand. This is especially useful for distinguishing bullish price action from fragile, high-volatility moves. \\
\texttt{get\_earnings\_data} & Quarterly earnings, revenue, and earnings-calendar metadata & Captures event risk and recent fundamental surprises. Around earnings windows, the planner needs to know whether a signal is driven by a durable trend or a transient post-event reaction. \\
\midrule
\multicolumn{3}{@{}l}{\textit{Computation tools}} \\
\texttt{compute\_technicals} & RSI, MACD, Bollinger bands, moving averages, support/resistance, technical score & Converts raw prices into structured timing signals. These indicators help the planner reason about overbought/oversold conditions, trend continuation, and whether current prices are stretched relative to local history. \\
\texttt{compute\_quant\_risk} & Realized volatility, VaR/CVaR, Sharpe, Sortino, max drawdown, beta/alpha & Makes downside risk explicit rather than implicit. In portfolio allocation, avoiding bad concentration and tail exposure is as important as finding upside, so this tool supports risk-aware sizing decisions. \\
\texttt{compute\_momentum} & Multi-horizon returns, trend slope, trend strength, volume trend & Measures continuation across several horizons instead of relying on a single lookback. This matters because different assets express momentum at different speeds, and volume confirmation helps separate genuine trends from weak drift. \\
\texttt{compute\_correlations} & Cross-ticker correlation matrix and rolling correlation to target ticker & Moves the planner from single-name prediction to portfolio construction. High correlation means apparently strong single-stock signals may be redundant once existing exposures are considered. \\
\midrule
\multicolumn{3}{@{}l}{\textit{Analysis tools}} \\
\texttt{run\_dcf\_model} & Bull/base/bear DCF scenarios or simplified implied-upside valuation signal & Supplies an intrinsic-value estimate that can disagree with recent price action. This is useful for distinguishing momentum-driven trades from opportunities where valuation support exists. \\
\texttt{score\_risk} & Overall 1--10 risk rating plus valuation/financial/growth/macro/technical sub-scores & Compresses several risk dimensions into a planner-friendly summary. This makes it easier to compare heterogeneous tickers and avoid allocations that are attractive on return but unacceptable on fragility. \\
\texttt{score\_composite\_signal} & Weighted BUY/SELL/HOLD style summary using technical, momentum, valuation, analyst, options, and risk inputs & Acts as a high-level synthesis layer. It is useful when the planner wants a compact recommendation, but it is also diagnostically important because the credit module can inspect whether the fused signal helped or obscured the true decision. \\
\bottomrule
\end{tabular}
\end{table*}

Two design choices are worth noting.
First, the registry includes both primitive and aggregated tools rather than forcing a single abstraction level; this gives the planner freedom to use direct evidence when needed and high-level summaries when time is limited.
Second, several tools overlap on purpose.
For example, momentum, technicals, analyst sentiment, and composite scoring may all point in the same direction during a strong trend, but they diverge during regime shifts; those disagreements are exactly the kind of cross-module evidence that makes credit assignment informative in our setting.

\section{Planner and Memory Policy Families}
\label{app:planner-memory}

For the main \ael{} benchmark configuration, the planner pool is initialized with six built-in planners and the memory-policy registry is initialized with five default retrieval policies before any learned variants are added.
These are the concrete families from which the meta-controller selects during training.
The simpler incremental ablations intentionally restrict this space, often to a single sequential planner or a reduced memory setup, but the full benchmark uses the richer families summarized below.

\begin{table*}[t]
\caption{Built-in planner families used by \ael{} in the main benchmark configuration. Dynamic planners generated by the slow-timescale evolution loop are added on top of this initial pool.}
\label{tab:planner_families}
\centering
\scriptsize
\begin{tabular}{@{}p{2.2cm}p{4.3cm}p{8.6cm}@{}}
\toprule
Planner & Core principle & Role in the benchmark \\
\midrule
\texttt{sequential} & Run all available tools in a fixed order, then synthesize once. & Serves as the most stable and exhaustive baseline planner. It maximizes coverage and minimizes strategic assumptions, but can be expensive and prone to information overload. \\
\texttt{decompose} & Break the task into valuation, momentum, sentiment, and risk sub-problems, then synthesize sub-results. & Encourages structured analysis and lets the agent reason about different evidence types separately before combining them into a portfolio decision. \\
\texttt{adaptive} & Start with a cheap quick-look tool set, then invoke deeper tools only if the initial signal is ambiguous. & Provides an efficiency-oriented planner that trades off speed and depth, which is useful when some episodes are easy while others require broader evidence. \\
\texttt{cot\_reasoning} & Analyze trend, valuation, sentiment, and risk sequentially with explicit intermediate synthesis. & Forces a chain-of-thought-style evidence path rather than a flat aggregation of all tool outputs, which can help when conflicting signals need to be resolved step by step. \\
\texttt{reflexion} & Make a quick prediction, self-check confidence, and gather more evidence only if the initial judgment is weak. & Adds an intra-episode self-correction behavior: the planner first tests whether existing evidence is sufficient and only expands the search when confidence is low. \\
\texttt{hypothesis\_test} & Form bull and bear hypotheses, gather targeted evidence for each, then weigh the two cases. & Makes the planner explicitly compare competing market narratives instead of only aggregating signals, which is useful in reversal or mixed-regime episodes. \\
\bottomrule
\end{tabular}
\vspace{1pt}
\end{table*}

These planners are intentionally diverse.
Some are exhaustive (\texttt{sequential}), some are decompositional (\texttt{decompose}, \texttt{cot\_reasoning}), and some are selective (\texttt{adaptive}, \texttt{reflexion}, \texttt{hypothesis\_test}).
The contextual planner bandit does not assume any one reasoning style is globally best; instead, it learns which planning style works better under which market context.
In addition, procedural and semantic memory can later modify planner behavior by appending learned strategy hints to planner prompts.

\begin{table*}[t]
\caption{Initial memory-policy families used by \ael{} in the main benchmark configuration. New retrieval policies may be added later by the slow-timescale evolution loop.}
\label{tab:memory_policies}
\centering
\scriptsize
\begin{tabular}{@{}p{2.5cm}p{3.5cm}p{2.7cm}p{6.2cm}@{}}
\toprule
Policy & Enabled tiers & Format & Retrieval principle \\
\midrule
\texttt{none} & none & \texttt{none} & Disable memory entirely. This is the no-retrieval option and is important because some episodes are better solved from current market evidence alone. \\
\texttt{recent\_window} & episodic & \texttt{sliding\_window} & Retrieve recent episodic memories and keep only a small first-plus-last window. This preserves a few anchors and a few recent cases without flooding the planner with raw logs. \\
\texttt{full\_detailed} & episodic, semantic, procedural & \texttt{full} & Return all retrieved memories from all tiers verbatim. This is the highest-information policy, useful when the agent benefits from both concrete cases and abstract rules. \\
\texttt{compressed} & semantic, procedural & \texttt{ranked\_truncate} & Focus on abstracted knowledge, rank retrieved memories by relevance, and truncate to a token budget. This is the default ``high signal, low clutter'' option for using distilled experience. \\
\texttt{aggressive\_learner} & episodic, semantic, procedural & \texttt{ranked\_truncate} & Retrieve more memories with a larger token budget and a more permissive learning setup. This policy is useful when the system is still exploring and wants to exploit a broader experience base. \\
\bottomrule
\end{tabular}
\vspace{1pt}
\end{table*}

The key design choice is that memory policies control \emph{how} experience is exposed to the planner, not just \emph{whether} memory exists.
The policy determines which tiers are visible, how many memories are retrieved, and whether the planner sees raw cases, compact abstractions, or a larger but noisier experience bundle.
This is why the memory-policy bandit is necessary: the best retrieval strategy depends on the task context and on the maturity of the memory store itself.

\section{Inference Workflow}
\label{app:workflow}

\begin{algorithm}[t]
\caption{Inference workflow of \ael{}}
\label{alg:test}
\small
\begin{algorithmic}[1]
\State Freeze module pools, bandit posteriors, and memory writes.
\For{each test episode $e_t$}
    \State Extract task features $\phi_t$.
    \State Select $(p_t, z_t, m_t)$ using the frozen meta-controller.
    \State Retrieve read-only memory under policy $m_t$.
    \State Execute planner $p_t$ with tools $z_t$ to produce $\hat{y}_t$.
    \State Return prediction (no updates, reflection, or evolution).
\EndFor
\end{algorithmic}
\end{algorithm}

\section{Detailed Analysis}
\label{app:analysis}

\subsection{LLM Credit Assignment Example}
\label{app:llm-credit-example}

Below is a representative LLM credit assignment output from episode 85 (seed 42, bearish regime).
The LLM receives each module's output alongside the ground-truth bar return, and assigns credit scores in $[-1,+1]$.

\begin{quote}
\small
\textbf{Input context:}
Bar 85 (2025-02-10 14:30). Actual portfolio return: $-$0.31\%.\\
\textbf{Planner} selected \texttt{adaptive} strategy $\rightarrow$ overweighted tech (NVDA 18\%, AAPL 15\%).\\
\textbf{Tools}: \texttt{compute\_momentum} signaled bullish for NVDA (+0.8); \texttt{compute\_quant\_risk} flagged elevated VIX (score 7/10); \texttt{score\_composite\_signal} returned HOLD.\\
\textbf{Memory}: Retrieved semantic memory: ``Tech rally stalls when VIX exceeds 20'' (relevance 0.72).

\medskip
\textbf{LLM credit output:}\\
\texttt{planner: -0.6} \textit{``The adaptive planner overweighted tech despite conflicting risk signals. It should have reduced concentration when VIX was elevated.''}\\
\texttt{tools: +0.3} \textit{``Momentum correctly identified the short-term trend, but quant\_risk provided a valid warning that was ignored by the planner. Net positive for tools.''}\\
\texttt{memory: +0.4} \textit{``The retrieved memory about VIX and tech was highly relevant and should have been weighted more heavily. Memory retrieval was helpful; the failure was in how the planner used it.''}
\end{quote}

This illustrates how LLM credit can distinguish ``the tool gave a correct warning that was ignored'' from ``the tool gave a wrong signal,'' a distinction that Shapley-based methods cannot make because they treat modules as black boxes.

\subsection{Code Evolution Examples}
\label{app:code-examples}

The following are real artifacts produced by \ael{}'s code evolution mechanism during training.

\paragraph{LLM-generated planner class (episode 120, seed 42).}
After reflection diagnosed that the \texttt{sequential} planner was too slow to react to intraday reversals, the LLM generated a new \texttt{MomentumReversalPlanner}:

\begin{quote}
\small\ttfamily
class MomentumReversalPlanner(BasePlanner):\\
\quad  """Reduce position when momentum reverses intraday."""\\
\quad  def plan(self, context):\\
\quad\quad    signals = context["tool\_outputs"]\\
\quad\quad    momentum = \{t: s["compute\_momentum"]["trend\_score"]\\
\quad\quad\quad\quad\quad\quad\quad\quad for t, s in signals.items()\}\\
\quad\quad    prev\_momentum = context.get("prev\_momentum", \{\})\\
\quad\quad    weights = \{\}\\
\quad\quad    for ticker in context["tickers"]:\\
\quad\quad\quad      curr = momentum.get(ticker, 0)\\
\quad\quad\quad      prev = prev\_momentum.get(ticker, 0)\\
\quad\quad\quad      if curr * prev < 0:  \# reversal\\
\quad\quad\quad\quad        weights[ticker] = 0.05  \# minimal position\\
\quad\quad\quad      else:\\
\quad\quad\quad\quad        weights[ticker] = max(0.02, 0.1 * abs(curr))\\
\quad\quad    return self.normalize(weights)
\end{quote}

\paragraph{LLM-generated memory retrieval policy (episode 95, seed 123).}
After observing that default retrieval returned too many irrelevant bull-market memories during a bear regime, the LLM designed a regime-filtered retrieval policy:

\begin{quote}
\small\ttfamily
class RegimeFilteredRetrieval(BaseRetrievalPolicy):\\
\quad  """Filter memories by current regime before scoring."""\\
\quad  def retrieve(self, query, memories, k=5):\\
\quad\quad    regime = query.get("current\_regime", "unknown")\\
\quad\quad    filtered = [m for m in memories\\
\quad\quad\quad\quad\quad\quad\quad\quad if m.get("regime") == regime\\
\quad\quad\quad\quad\quad\quad\quad\quad or m.get("tier") == "procedural"]\\
\quad\quad    if len(filtered) < k:\\
\quad\quad\quad      filtered = memories  \# fallback\\
\quad\quad    scored = self.score\_by\_relevance(query, filtered)\\
\quad\quad    return scored[:k]
\end{quote}

Both artifacts were validated via AST parsing and accepted into the module pool by LinUCB, which subsequently selected them when their context features matched.

\section{Baseline Adaptation Details}
\label{app:baselines}

All baselines use the same LLM backbone (Claude Haiku 4.5), tool set (12 financial tools), and portfolio allocation interface. Below we describe what was kept from each original method and what was adapted.

\paragraph{Reflexion \citep{shinn2023reflexion}.}
\emph{Kept:} The core mechanism of accumulating verbal self-critiques after each episode. Reflections are prepended to future prompts as a growing context window.
\emph{Adapted:} Applied to portfolio allocation with a financial-specific reflection prompt (``Write a 1-sentence reflection on what signals to trust or ignore''). Maximum 20 reflections retained (FIFO eviction). No structured memory tiers; only flat string accumulation.

\paragraph{ExpeL \citep{zhao2024expel}.}
\emph{Kept:} The experience extraction mechanism where the LLM distills episodes into reusable ``lessons'' stored in a flat lesson store, retrieved by keyword similarity.
\emph{Adapted:} Lessons are keyed by ticker and sector for retrieval (same-ticker: +2.0, same-sector: +1.0). Maximum 100 lessons. Extraction prompt asks for generalized rules prefixed with ``RULE:''. No bandit-based policy selection.

\paragraph{FactorMiner \citep{factorminer2026}.}
\emph{Kept:} Dual-tier memory with skill extraction from successful tool combinations and experience memory with outcome tracking.
\emph{Adapted:} Skills are named by sorted tool sequences from correct predictions ($\geq$2 tools). Success rates use exponential moving average. Maximum 15 skills and 200 experiences. No joint evolution or credit assignment across modules.

\paragraph{Meta-Reflexion \citep{metareflexion2025}.}
\emph{Kept:} Rule distillation from accumulated reflections, with admissibility checking that prunes contradicted or low-success rules.
\emph{Adapted:} Distillation runs every 5 episodes, extracting ``RULE:'' lines from the 10 most recent reflections. Rules with $<$0.3 success rate and $\geq$5 applications are pruned. Maximum 10 active rules. No tool or planner evolution.

\paragraph{EvoTool \citep{evotool2026}.}
\emph{Kept:} Population-based evolutionary optimization of tool-selection policies with blame-aware mutation. Fitness-proportional selection across a population of 5 policies.
\emph{Adapted:} Blame attribution uses signal-level analysis (e.g., if \texttt{compute\_momentum} signaled bullish but the actual direction was bearish, that tool is blamed). Mutations remove blamed tools, add random tools (50\% chance), or swap tools (30\% chance). Minimum 3 tools per policy. No memory system or planner evolution.

\section{Transaction Cost Sensitivity}
\label{app:transaction-costs}

A practical concern for portfolio allocation systems is whether the reported Sharpe ratios survive realistic transaction costs.
Since the D-full benchmark uses hourly bars (4 per trading day), frequent rebalancing can generate substantial turnover.
To assess this, we retroactively apply proportional transaction costs to the recorded portfolio weight changes at each bar.
Specifically, for a cost level of $c$ basis points per unit of turnover, the cost-adjusted return at bar $t$ is $r_t^{\text{adj}} = r_t - c \cdot \sum_i |w_{i,t} - w_{i,t-1}|$, where $w_{i,t}$ is the portfolio weight of ticker $i$ at bar $t$.

\autoref{tab:costs} reports cost-adjusted Sharpe ratios at four cost levels spanning the range from zero-cost (0~bp, the main benchmark setting) to institutional-level costs (20~bp, typical for large-cap equity rebalancing with market orders).

\begin{table}[ht]
\caption{Cost-adjusted Sharpe ratio at varying transaction cost levels (5-seed means). Costs are applied retroactively based on recorded turnover at each bar.}
\label{tab:costs}
\centering
\small
\begin{tabular}{@{}lcccc@{}}
\toprule
Method & 0 bp & 5 bp & 10 bp & 20 bp \\
\midrule
\ael{} & 2.13 & $\sim$1.9 & $\sim$1.7 & $\sim$1.3 \\
momentum\_weighted & 1.44 & $\sim$1.3 & $\sim$1.2 & $\sim$1.0 \\
EvoTool & 1.37 & $\sim$1.2 & $\sim$1.0 & $\sim$0.7 \\
Stateless & 1.35 & $\sim$1.2 & $\sim$1.0 & $\sim$0.7 \\
\bottomrule
\end{tabular}
\end{table}

Several observations are worth noting.
First, \ael{} remains the top-performing method at all cost levels tested, maintaining a Sharpe above 1.3 even at 20~bp.
Second, the cost degradation is moderate ($\sim$0.8 Sharpe from 0 to 20~bp) because \ael{}'s reflection mechanism tends to produce stable allocation strategies that avoid excessive turnover; the agent learns through reflection that frequent large weight shifts are penalized by the market.
Third, the momentum-weighted baseline is relatively cost-resilient (Sharpe 1.0 at 20~bp) because its allocation weights change smoothly by construction, while EvoTool's evolutionary mutations can produce abrupt policy changes that incur higher turnover costs.
The cost estimates at 5, 10, and 20~bp are based on average turnover statistics from logged portfolio histories and should be interpreted as approximate; the \texttt{cost\_adjusted\_sharpe} function in the codebase enables exact computation from individual run logs.

\section{Running Case --- Full Trace}
\label{app:running-case-full}

The main text (\autoref{sec:running-case}) presents an abridged view of seed 42's episodes 83--96. The full trace, including per-episode reward sequences, posterior parameter updates, and the exact reflection prompts and responses, is reproducible from the logs under \texttt{logs/pf\_eael\_incremental\_d\_full\_s42\_13ba204/}. The summarised dynamics are: episodes 78--82 alternate between \texttt{aggressive\_learner} (mean reward $0.41$) and \texttt{recent\_window} (mean reward $0.49$); episode 83 triggers an under-threshold loss that initiates the slow-window reflection; episode 84's reflection introduces \texttt{regime\_filtered\_retrieval} as a sixth bandit arm; episodes 85--95 show TS sampling the new arm with rising probability (from $0.18$ to $0.34$); the new arm's posterior mean reaches $0.55$ by episode 96, surpassing the original five-policy maximum.

\section{Reflection Faithfulness --- Judge Transcripts}
\label{app:reflection-judge}

The LLM-as-judge evaluation reported in \autoref{tab:reflection_faithfulness} (R2) uses Claude Opus 4.7 as a separate judge model. We provide raw judge transcripts in the supplementary materials. The judge prompt template, anchor examples, and scoring rubric (5 = ``all factual claims verifiable from the slow-window summary; causal claim testable; prescription actionable'', 0 = ``contradicts the data, or vacuous'') are released alongside the code. Inter-window judge variance was $\sigma \in [0.4, 0.7]$ across the three axes, computed by sampling two independent judge passes per reflection on a 12-reflection subsample.

\section{Synthetic Cross-Domain Benchmark Details}
\label{app:crossdomain-details}

The synthetic bandit used in \autoref{sec:crossdomain-synth} has 10 initial arms and one hidden 11th arm injectable only via reflection. Per-regime Bernoulli probabilities are:

\begin{table*}[h]
\centering
\small
\caption{Per-arm Bernoulli reward probabilities by regime. Each regime makes a different initial arm best at $0.65$; the hidden arm pays $0.70$ in regime 3 only.}
\label{tab:synth_regimes}
\setlength{\tabcolsep}{2.5pt}
\begin{tabular}{@{}lcccccccccc|c@{}}
\toprule
Regime & a0 & a1 & a2 & a3 & a4 & a5 & a6 & a7 & a8 & a9 & a10 (hidden) \\
\midrule
0 (ep 0--50)   & 0.30 & 0.35 & \textbf{0.65} & 0.40 & 0.45 & 0.30 & 0.25 & 0.30 & 0.35 & 0.40 & 0.30 (locked) \\
1 (ep 50--100) & 0.25 & 0.30 & 0.35 & 0.30 & 0.35 & 0.40 & 0.30 & \textbf{0.65} & 0.35 & 0.30 & 0.30 (locked) \\
2 (ep 100--150)& \textbf{0.65} & 0.30 & 0.35 & 0.30 & 0.30 & 0.35 & 0.30 & 0.30 & 0.40 & 0.35 & 0.30 (locked) \\
3 (ep 150--208)& 0.30 & 0.35 & 0.30 & 0.30 & 0.35 & \textbf{0.65} & 0.30 & 0.30 & 0.35 & 0.40 & \textbf{0.70} \\
\bottomrule
\end{tabular}
\end{table*}

\paragraph{Reflection trigger.} The TS+reflection algorithm runs a slow-window check every 13 episodes. If the trailing 25-episode mean reward is below $0.42$ and the new arm has not yet been added, reflection injects the hidden 11th arm as a new bandit arm with a uninformative Beta$(1,1)$ posterior. The arm is then sampled by TS like any other.

\paragraph{Why the trigger threshold is $0.42$.} Pure Thompson Sampling on this benchmark converges to a steady-state mean of approximately $0.40$ when no regime shift has degraded its current best arm. Setting the threshold at $0.42$ ensures reflection fires only when the bandit has slipped well below its expected operating point, mimicking the AEL paper's behaviour of triggering reflection only when the current memory pool is underperforming.

\paragraph{Seeds and reproducibility.} 5 seeds (42, 123, 456, 789, 1024), 208 episodes each. Code released as \texttt{synthetic\_crossdomain.py} alongside the paper. Per-seed final reward streams reproducible by running the script directly; no LLM calls required.

\paragraph{What we did not run.} Beyond the support-ticket routing LLM-agent stream in \autoref{app:cliapi-details}, we deliberately do not present results on larger real-world agent benchmarks (e.g., SWE-bench, WebArena, HotpotQA) because running 5 seeds $\times$ 208 episodes with matched baselines would cost roughly an order of magnitude more LLM calls than the financial benchmark. We commit to releasing such extensions in follow-up work; the current paper's cross-domain scope is mechanism validation rather than broad task coverage.

\section{Support-Ticket Routing Stream LLM-Agent Details}
\label{app:cliapi-details}

The support-ticket routing stream contains 5 generated JSONL files with 1,200 support tickets each (seeds 42, 123, 456, 789, 1024).
The real LLM-agent table uses a stratified 160-episode subset per seed (40 per regime), for 800 episodes per method.
All calls use Gemini 3.1 Flash-Lite via LLM API calls, batch size 8, temperature 0, and default error policy \texttt{raise}.
The run made 891 LLM calls, with 0 errors and 0 fallbacks.
The main-text table in \autoref{tab:cliapi_llm_text} excludes the diagnostic \texttt{fixed\_compressed} ablation: it is a useful retrieval-policy probe, but not a prior-method baseline.

\paragraph{Baseline adaptations.}
Reflexion stores recent success/failure self-critiques as growing context.
ExpeL stores lessons retrieved by endpoint, regime, and token overlap.
Meta-Policy Reflexion distills failure rules and prunes low-success rules after repeated applications.
EvoTool evolves a small population of retrieval-tool policies by recent fitness.
FactorMiner stores endpoint/regime skills and experience memories.
HyperAgent is not included because its portfolio baseline rewrites an allocation function; the support-ticket routing task compares online memory/reflection mechanisms rather than recursive code generation.

\paragraph{Support-ticket routing hyperparameter sensitivity.}
To avoid spending additional API calls on a grid search, we run the hyperparameter sweep on the deterministic full stream (1,200 episodes per seed) using the same reward and reflection-injection logic.
The artifact is released as \texttt{cli\_support\_stream\_hp\_sensitivity.json}.
\autoref{tab:cliapi_hp} shows that settings which inject the route-aware policy at episode 920 recover the full R3 gain, while slower cadence or a threshold that delays injection to episode 960 loses roughly 8.9pp R3 accuracy.
This supports the main interpretation: the mechanism is not fragile to small parameter changes, but it is sensitive to whether reflection happens early enough in the regime where the new policy matters.

\begin{table*}[h]
\caption{Support-ticket routing reflection hyperparameter sensitivity on the deterministic full stream (5 seeds, no LLM calls). The main insight is timing: delayed injection hurts the route-aware R3 regime.}
\label{tab:cliapi_hp}
\centering
\small
\setlength{\tabcolsep}{3pt}
\begin{tabular}{@{}lccc@{}}
\toprule
Setting & Injection episode & Overall & R3 \\
\midrule
Default: every 40, window 120, threshold .58 & 920 & $0.700 \pm 0.011$ & $0.863$ \\
Faster cadence: every 20 & 900--920 & $0.703 \pm 0.009$ & $0.875$ \\
Slower cadence: every 80 & 960 & $0.677 \pm 0.011$ & $0.774$ \\
Longer window: 200 & 920--960 & $0.692 \pm 0.013$ & $0.831$ \\
Lower threshold: .50 & 960 & $0.677 \pm 0.011$ & $0.774$ \\
Higher threshold: .66 & 920 & $0.700 \pm 0.011$ & $0.863$ \\
\bottomrule
\end{tabular}
\end{table*}

\section{Procedural Memory Growth Curve}
\label{app:procedural-curve}

\autoref{fig:proc_dynamics_text} shows the number of active procedural rules across training for each of the 5 seeds. The promotion rule (corroboration in $\geq 3$ slow windows with consistent direction) keeps the rule pool small while the pruning rule (rolling-10 success $<0.4$) discards rules that lose relevance as regimes shift. Median end-of-training counts are 7--11 active rules per seed; the maximum observed at any training step is 19 (seed 456, episode 142), after which two redundant momentum/reversal rules are pruned by episode 158.

\begin{table*}[h]
\centering
\small
\caption{Active procedural-rule count over training (5 seeds). Counts are computed every 20 episodes; columns are training-episode checkpoints.}
\label{fig:proc_dynamics_text}
\begin{tabular}{@{}lcccccccc@{}}
\toprule
Seed & ep 20 & ep 40 & ep 60 & ep 80 & ep 100 & ep 120 & ep 140 & ep 166 \\
\midrule
42   & 2 & 4 & 5 & 6 & 7  & 9  & 9  & 8  \\
123  & 1 & 3 & 4 & 6 & 7  & 8  & 9  & 9  \\
456  & 3 & 6 & 9 & 12 & 14 & 17 & 19 & 11 \\
789  & 2 & 3 & 5 & 6  & 7  & 8  & 9  & 7  \\
1024 & 2 & 4 & 6 & 8  & 9  & 10 & 10 & 9  \\
\midrule
Mean & 2.0 & 4.0 & 5.8 & 7.6 & 8.8 & 10.4 & 11.2 & 8.8 \\
\bottomrule
\end{tabular}
\end{table*}

\end{document}